\documentclass{article} 
\usepackage{iclr2024_conference,times}

\usepackage{amsmath,amsfonts,bm}

\def\Secref#1{Section~\ref{#1}}

\def\eqref#1{equation~\ref{#1}}

\def\1{\bm{1}}

\def\vt{{\bm{t}}}

\def\vw{{\bm{w}}}
\def\vx{{\bm{x}}}

\def\vz{{\bm{z}}}

\DeclareMathAlphabet{\mathsfit}{\encodingdefault}{\sfdefault}{m}{sl}
\SetMathAlphabet{\mathsfit}{bold}{\encodingdefault}{\sfdefault}{bx}{n}

\newcommand{\E}{\mathbb{E}}

\newcommand{\R}{\mathbb{R}}

\newcommand{\Var}{\mathrm{Var}}

\newcommand{\Cov}{\mathrm{Cov}}

\usepackage{graphicx} 

\usepackage{subcaption}
\usepackage{hyperref}
\usepackage{url}
\usepackage{algorithm}
\usepackage{algorithmicx}
\usepackage{algpseudocode}
\usepackage{float}%
\usepackage{booktabs}%

\usepackage{wrapfig}
\usepackage{multirow}

\usepackage[capitalize,noabbrev]{cleveref}

\title{BayesDiff: Estimating Pixel-wise Uncertainty in Diffusion via Bayesian Inference}

\author{Siqi Kou\textsuperscript{1},  Lei Gan\textsuperscript{4},  Dequan Wang\textsuperscript{1,5},  Chongxuan Li\textsuperscript{2,3}\thanks{Corresponding authors.}\;\,, and Zhijie Deng\textsuperscript{1}\footnotemark[1]  \\
\textsuperscript{1}Qing Yuan Research Institute, SEIEE, Shanghai Jiao Tong University \\
\textsuperscript{2}Gaoling School of Artificial Intelligence, Renmin University of China \\ \textsuperscript{3}Beijing Key Laboratory of Big Data Management and Analysis Methods \\
\textsuperscript{4}School of Computer Science, Fudan University \textsuperscript{5}Shanghai Artificial Intelligence Laboratory \\
\texttt{\small happy-karry@sjtu.edu.cn, 21307130211@m.fudan.edu.cn} \\
\texttt{\small dequanwang@sjtu.edu.cn, chongxuanli1991@gmail.com, zhijied@sjtu.edu.cn}
}

\iclrfinalcopy
\begin{document}

\maketitle

\begin{abstract}
Diffusion models have impressive image generation capability, but low-quality generations still exist, and their identification remains challenging due to the lack of a proper sample-wise metric. To address this, we propose \emph{BayesDiff}, a pixel-wise uncertainty estimator for generations from diffusion models based on Bayesian inference. In particular, we derive a novel uncertainty iteration principle to characterize the uncertainty dynamics in diffusion, and leverage the last-layer Laplace approximation for efficient Bayesian inference.
The estimated pixel-wise uncertainty can not only be aggregated into a sample-wise metric to filter out low-fidelity images but also aids in augmenting successful generations and rectifying artifacts in failed generations in text-to-image tasks. Extensive experiments demonstrate the efficacy of BayesDiff and its promise for practical applications. Our code is available at \url{https://github.com/karrykkk/BayesDiff}.
\end{abstract}

\section{Introduction}

The ability of diffusion models to gradually denoise noise vectors into natural images has paved the way for numerous applications, including image synthesis~\citep{dhariwal2021diffusion,rombach2022high}, image inpainting~\citep{lugmayr2022repaint}, text-to-image generation~\citep{saharia2022photorealistic,gu2022vector,zhang2023text}, etc.
However, there are still inevitable low-quality generations causing poor user experience in downstream applications.
A viable remediation is to filter out low-quality generations, which, yet, cannot be trivially realized due to the lack of a proper metric for image quality identification. 
For example, the traditional metrics such as the Fréchet Inception Distance (FID)~\citep{heusel2017gans} and Inception Score (IS)~\citep{salimans2016improved} scores estimate the distributional properties of the generations instead of sample-wise quality. 

Bayesian uncertainty has long been used to identify data far from the manifold of training samples~\citep{maddox2019simple,deng2021libre}. 
The notion is intuitive---the Bayesian posterior delivers low uncertainty for the data witnessed during training while high uncertainty for the others. 
This fits the requirement that the generations returned to the users should be as realistic as the training images. 
However, the integration of Bayesian uncertainty and diffusion models is not straightforward. Diffusion models typically involve large-scale networks, necessitating efficient Bayesian inference strategies. 
Additionally, the generation of images typically involves an intricate reverse diffusion process, which adds to the challenge of accurately quantifying their uncertainty.

\begin{figure}[t]
\begin{center}
\includegraphics[width=.7\textwidth]{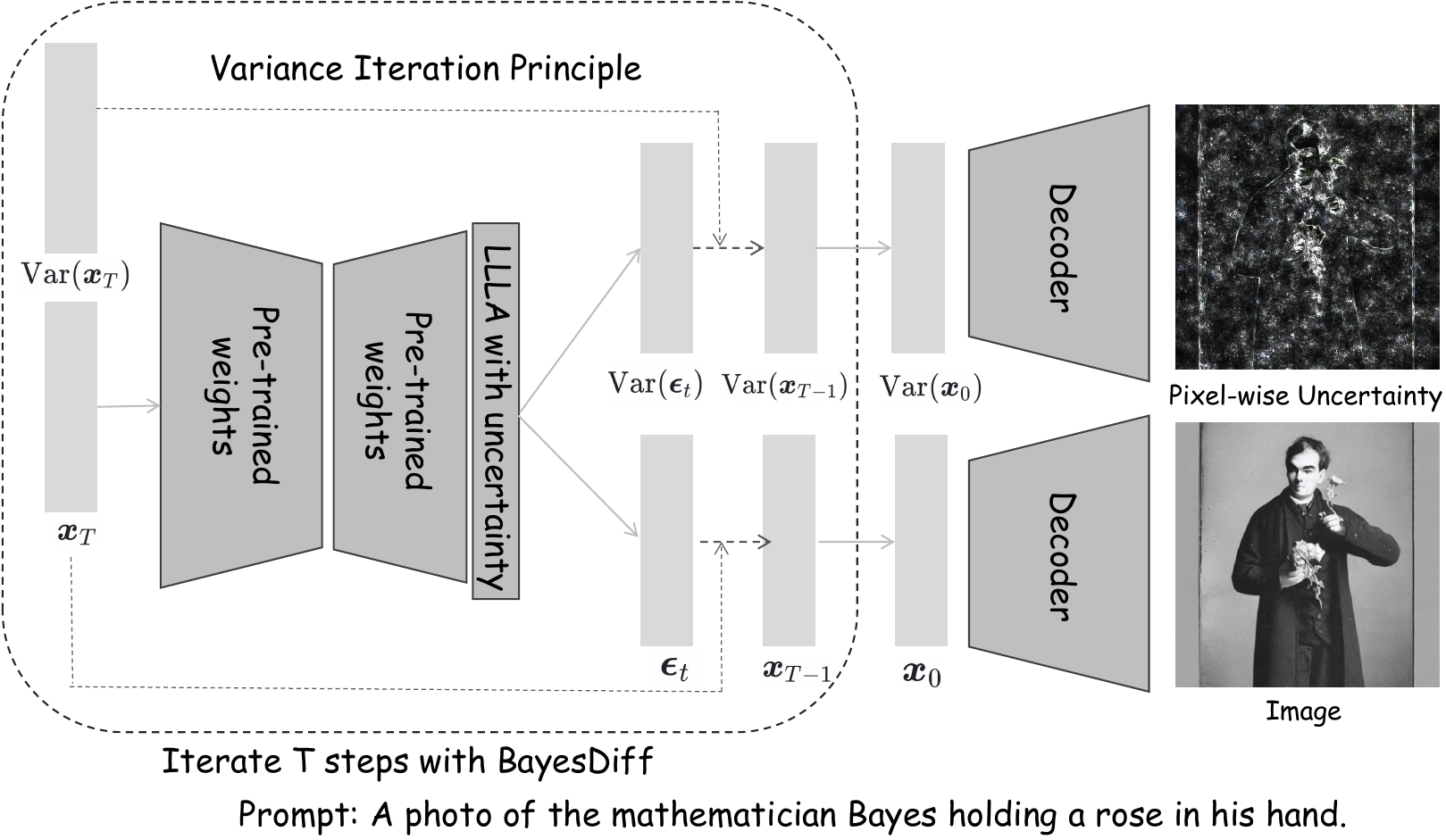}
\end{center}
\caption{Given an initial point $\vx_T \sim \mathcal{N}(\mathbf{0}, \mathbf{I})$, our BayesDiff framework incorporates uncertainty into the denoising process and generates images with pixel-wise uncertainty estimates.}
\label{fig:intro}
\end{figure}

To address these challenges, we propose BayesDiff, a framework for estimating the pixel-wise Bayesian uncertainty of the images generated by diffusion models. %
We develop a novel uncertainty iteration principle that applies to various sampling methods to characterize the dynamics of pixel-wise uncertainty in the reverse diffusion process, as illustrated in Figure~\ref{fig:intro}. 
We leverage the last-layer Laplace approximation (LLLA)~\citep{kristiadi2020being,daxberger2021laplace} for efficient Bayesian inference of pre-trained score models.
Finally, BayesDiff enables the simultaneous delivery of image samples and pixel-wise uncertainty estimates. %
The naive BayesDiff can be slow due to the involved Monte Carlo estimation, so we further develop an accelerated variant of it to enhance efficiency.

We can aggregate the obtained pixel-wise uncertainty into image-wise metrics (e.g., through summation) for generation filtering. 
Through extensive experiments conducted on ADM~\citep{dhariwal2021diffusion}, U-ViT~\citep{bao2023all}, and Stable Diffusion~\citep{rombach2022high} using samplers including DDIM~\citep{song2020denoising} and DPM-Solver~\citep{lu2022dpm}, we demonstrate the efficacy of BayesDiff in filtering out images with cluttered backgrounds. 
Moreover, in text-to-image generation tasks, we surprisingly find that pixel-wise uncertainty can enhance generation diversity by augmenting good generations, and rectify failed generations containing artifacts and mismatching with textual descriptions. 
Additionally, we perform comprehensive ablation studies %
to seek a thorough and intuitive understanding of the estimated pixel-wise uncertainty.

\section{Background}

This section briefly reviews the methodology of diffusion probabilistic models and introduces Laplace approximation (LA), a classic method for approximate Bayesian inference.

\subsection{Diffusion Models}
\label{sec:diffusion}
Let $\vx \in \R^{c\times h \times w}$ denote an image. 
A diffusion model (DM)~\citep{ho2020denoising} typically assumes a forward process gradually diffusing data distribution $q(\vx)$ towards $q_t(\vx_t), \forall t \in [0, T]$, with $q_T(\vx_T) = \mathcal{N}(\mathbf{0}, \tilde{\sigma}^2\mathbf{I})$ as a trivial Gaussian distribution. %
The transition distribution obeys a Gaussian formulation, i.e., $q_t(\vx_t|\vx)=\mathcal{N}(\vx_t; \alpha_t\vx, \sigma^2_t \mathbf{I})$, where $\alpha_t, \sigma_t \in \mathbb{R}^+$. 
The reverse process is defined with the data score $\nabla_{\vx_t} \log q_t(\vx_t)$, which is usually approximated by $-{\epsilon}_{\theta}(\vx_t, t) / \sigma_t$ with ${\epsilon}_{\theta}$ as a parameterized noise prediction network trained by minimizing:
\begin{equation}
\label{eq:loss}
    \mathbb{E}_{\vx \sim q(\vx),  \boldsymbol{\epsilon} \sim \mathcal{N}(\mathbf{0}, \mathbf{I}), t\sim \mathcal{U}(0, T)} [w(t) \Vert {\epsilon}_{\theta}(\alpha_t \vx + \sigma_t \boldsymbol{\epsilon}, t) - \boldsymbol{\epsilon} \Vert_2^2]
\end{equation}
where $w(t)$ denotes a weighting function. 

\citet{kingma2021variational} show that the stochastic differential equation (SDE) satisfying the transition distribution specified above takes the form of
\begin{equation}
    d \vx_t = f(t)\vx_t dt + g(t) d \boldsymbol{\omega}_t,
\end{equation}
where $\boldsymbol{\omega}_t$ is the standard Wiener process, $f(t) = \frac{d \log \alpha_t}{d t}$, and $g(t)^2 = \frac{d \sigma^2_t}{d t} - 2 \frac{d \log \alpha_t}{d t}\sigma^2_t$. 
An SDE and an ordinary differential equation (ODE) starting from $\vx_T$ capturing the reverse process can be constructed as~\citep{songscore}:
\begin{equation}
\label{eq:sde}
    \begin{aligned}
        d\vx_t = [f(t)\vx_t + \frac{g(t)^2 }{\sigma_t}\boldsymbol{\epsilon}_t] dt + g(t)d \bar{\boldsymbol{\omega}}_t
    \end{aligned}
\end{equation}
\begin{equation}
\label{eq:ode}
    \frac{d\vx_t}{dt} = f(t)\vx_t + \frac{g(t)^2}{2\sigma_t} \boldsymbol{\epsilon}_t,
\end{equation}
where $\bar{\boldsymbol{\omega}}_t$ is the reverse-time Wiener process when time flows backward from $T$ to $0$, $d t$ is an infinitesimal negative timestep and $\boldsymbol{\epsilon}_t := {\epsilon}_{\theta}(\vx_t, t)$ denotes the noise estimated by the neural network (NN) model. We could sample $\vx_0\sim q(\vx)$ from $\vx_T \sim q_T(\vx_T )$ by running backwards in time with the numerical solvers~\citep{lu2022dpm,karras2022elucidating} for \cref{eq:sde} or \cref{eq:ode}.

\subsection{Bayesian Inference in Deep Models and Laplace Approximation}
\label{sec:la}

Bayesian inference turns a deterministic neural network into a Bayesian neural network (BNN). 
Let $p(\mathcal{D}| \theta)$ denote the data likelihood of the NN $f_\theta$ for the dataset $\mathcal{D}=\{(\vx^{(n)}, y^{(n)})\}_{n=1}^N$. 
Assuming an isotropic Gaussian prior $p(\theta)$, %
BNN methods estimate the Bayesian posterior $p(\theta|\mathcal{D}) = p(\theta) p(\mathcal{D}|\theta) / p(\mathcal{D})$, where $p(\mathcal{D}|\theta):= \prod_{n} p(y^{(n)}|\vx^{(n)},\theta)= \prod_{n} p(y^{(n)}|f_\theta(\vx^{(n)}))$, and predict for new data $\vx^*$ with $p(y|\vx^*, \mathcal{D}) = \E_{p(\theta|\mathcal{D})} p(y|f_\theta(\vx^*))$. 

Due to NNs' high nonlinearity, analytically computing $p(\theta|\mathcal{D})$ is often infeasible. 
Hence, approximate inference techniques such as variational inference (VI)~\citep{blundell2015weight,hernandez2015probabilistic,louizos2016structured,zhang2018noisy,khan2018fast}, Laplace approximation (LA)~\citep{mackay1992bayesian,ritter2018scalable}, Markov chain Monte Carlo (MCMC)~\citep{welling2011bayesian,chen2014stochastic,zhang2019cyclical}, and particle-optimization based variational inference (POVI)~\citep{liu2016stein} are routinely introduced to yield an approximation $q(\theta)\approx p(\theta|\mathcal{D})$. 
Among them, LA has recently gained particular attention because it can apply to pre-trained models effortlessly in a post-processing manner and enjoy strong uncertainty quantification (UQ) performance~\citep{foong2019between,daxberger2021laplace,daxberger2021bayesian}. 

LA approximates $p(\theta|\mathcal{D})$ with $q(\theta)=\mathcal{N}(\theta ; \theta_{\text{MAP}}, \boldsymbol{\Sigma})$, where $\theta_{\text{MAP}}$ denotes the maximum a posteriori (MAP) solution $\theta_{\text{MAP}} =\arg \max _{\theta} \log p(\mathcal{D}| \theta)+\log p(\theta)$, and 
$\boldsymbol{\Sigma}=[-\nabla_{\theta \theta}^2(\log p(\mathcal{D}| \theta)+\log p(\theta))|_{\theta=\theta_{\text{MAP}}}]^{-1}$ characterizes the Bayesian uncertainty over model parameters. 
Last-layer Laplace approximation (LLLA)~\citep{kristiadi2020being,daxberger2021laplace} further improves the efficiency of LA by concerning only the parameters of the last layer of the NN. 
It is particularly suited to the problem of uncertainty quantification for DMs, as DMs are usually large.

\section{Methodology}
We incorporate LLLA into the noise prediction model in DMs for uncertainty quantification at a single timestep. 
We then develop a novel algorithm to estimate the dynamics of pixel-wise uncertainty along the reverse diffusion process. 
We also develop a variant of it for practical acceleration.

\subsection{Laplace Approximation on Noise Prediction Model}

\label{sec:la_score}

Usually, the noise prediction model is trained to minimize \cref{eq:loss} under a weight decay regularizer, which corresponds to the Gaussian prior on the NN parameters. 
Namely, we can regard the pre-trained parameters as a MAP estimation and perform LLLA. 
Of note, the noise prediction problem corresponds to a regression under Gaussian likelihood, based on which we estimate the Hessian matrix involved in the approximate posterior (or its variants such as the generalized Gauss-Newton matrix). 
The last layer in DMs is often linear w.r.t. the parameters, so the Gaussian approximate posterior distribution on the parameters directly leads to a Gaussian posterior predictive:
\begin{equation}
\label{eq:post-pred}
    p(\boldsymbol{\epsilon}_t |\vx_t, t, \mathcal{D}) \approx \mathcal{N}({\epsilon}_{\theta}(\vx_t, t), \mathrm{diag}({\gamma}^2_\theta(\vx_t, t))),
\end{equation}
where we abuse $\theta$ to denote the parameters of the pre-trained DM.
We keep only the diagonal elements in the Gaussian covariance, ${\gamma}^2_\theta(\vx_{t}, t)$, because they refer to the pixel-wise variance of the predicted noise, i.e., the pixel-wise prediction uncertainty of $\boldsymbol{\epsilon}_t$. Implementation details regarding the LLLA are shown in Appendix~\ref{llla_implement_details}.

\subsection{Pixel-wise Uncertainty Estimation in Reverse Denoising Process}
\label{sec:uq_discrete}

Next, we elaborate on integrating the uncertainty obtained above into the reverse diffusion process. 

Although various sampling methods may correspond to various reverse diffusion processes, the paradigm for characterizing the uncertainty dynamics is similar. 
Take the SDE-form one (\cref{eq:sde}) for example, introducing Bayesian uncertainty to the noise prediction model yields:
\begin{equation}
    d\vx_t = [f(t)\vx_t + \frac{g(t)^2 }{\sigma_t}\boldsymbol{\epsilon}_t] dt + g(t)d \bar{\boldsymbol{\omega}}_t, 
    \label{eq:sampling}
\end{equation}
where $\boldsymbol{\epsilon}_t \sim \mathcal{N}({\epsilon}_{\theta}(\vx_t, t), \mathrm{diag}({\gamma}^2_\theta(\vx_t, t)))$. 
Assume the following discretization for it:
\begin{equation}
\label{eq:disc}
\small
\vx_{t-1}=\vx_{t}-(f(t)\vx_t+\frac{g(t)^2}{\sigma_t}\boldsymbol{\epsilon}_t)+g(t)\boldsymbol{z},
\end{equation}
where $\boldsymbol{z} \sim \mathcal{N}(\mathbf{0}, \mathbf{I})$. 
To estimate the pixel-wise uncertainty of $\vx_{t-1}$, we apply variance estimation to both sides of the equation, giving rise to
\begin{equation}
\label{eq:var-ite}
    \Var(\vx_{t-1})=(1-f(t))^2\Var(\vx_t)-(1-f(t))\frac{g(t)^2}{\sigma_t}\Cov(\vx_t,\boldsymbol{\epsilon}_t)+\frac{g(t)^4}{\sigma_t^2}\Var(\boldsymbol{\epsilon}_t)+g(t)^2\mathbf{1},
\end{equation}
where $\Cov(\vx_t,\boldsymbol{\epsilon}_t) \in \R^{c\times w \times h}$ denotes the pixel-wise covariance between $\vx_t$ and $\boldsymbol{\epsilon}_t$. %
With this, we can iterate over it to estimate the pixel-wise uncertainty of the final $\vx_0$. 
Recalling that $\Var(\boldsymbol{\epsilon}_t) = {\gamma}^2_\theta(\vx_t, t)$, 
the main challenges then boils down to estimating $\Cov(\vx_t,\boldsymbol{\epsilon}_t)$.

\textbf{The estimation of $\Cov(\vx_t,\boldsymbol{\epsilon}_t)$.} By the law of total expectation $\E(\E(X|Y))=\E(X)$, there is
\begin{equation}
\begin{aligned}
\Cov(\vx_t,\boldsymbol{\epsilon}_t)
&= \E(\vx_t\odot \boldsymbol{\epsilon}_t)-\E\vx_t \odot \E\boldsymbol{\epsilon}_t \\
&= \E_{\vx_t}(\E_{\boldsymbol{\epsilon}_t|\vx_t}(\vx_t\odot \boldsymbol{\epsilon}_t|\vx_t))-\E\vx_t \odot \E_{\vx_t}(\E_{\boldsymbol{\epsilon}_t|\vx_t} (\boldsymbol{\epsilon}_t|\vx_t)) \\
&= \E_{\vx_t}(\vx_t\odot {\epsilon}_{\theta}(\vx_t, t))-
\E\vx_t \odot \E_{\vx_t}({\epsilon}_{\theta}(\vx_t, t))
\end{aligned}
\end{equation}
where $\odot$ denotes the element-wise multiplication.
To estimate this, we need the distribution of $\vx_t$.

We notice that it is straightforward to estimate $\E(\vx_t)$ via a similar iteration rule to \cref{eq:var-ite}:
\begin{equation}
\label{eq:exp-sde}
    \E(\vx_{t-1})= (1-f(t))\E(\vx_t)-\frac{g(t)^2}{\sigma_t}\E(\boldsymbol{\epsilon}_t).
\end{equation}
Given these, we can reasonably assume $\vx_t$ follows $\mathcal{N}(\E(\vx_t), \Var(\vx_t))$, and then $\Cov(\vx_t,\boldsymbol{\epsilon}_t)$ can be approximated with Monte Carlo (MC) estimation:
\begin{equation}
\label{eq:mc}
\Cov(\vx_t,\boldsymbol{\epsilon}_t) \approx \frac{1}{S}\sum_{i=1}^S(\vx_{t,i}\odot {\epsilon}_{\theta}(\vx_{t,i}, t))- \E\vx_t \odot \frac{1}{S}\sum_{i=1}^S{\epsilon}_{\theta}(\vx_{t,i}, t),
\end{equation}
where $\vx_{t,i} \sim \mathcal{N}(\E(\vx_t), \Var(\vx_t)), i=1,\dots,S$. 

\textbf{Applying our method to existing samplers.} The derivation from \cref{eq:disc} to \cref{eq:mc} presents our general methodology based on the classical Euler sampler of reverse-time SDE, which can be applied to an arbitrary existing sampler of diffusion models in principle. For broader interests and simplicity, we show the explicit rules for DDPM~\citep{ho2020denoising} as an instance. %

Specifically, the sampling rule of DDPM~\citep{ho2020denoising} is:
\begin{equation}
\mathbf{x}_{t-1}=\frac{1}{\sqrt{\alpha_t^{\prime}}}(\mathbf{x}_t-\frac{1-\alpha_t^{\prime}}{\sqrt{1-\bar{\alpha}^{\prime}_t}}\boldsymbol{\epsilon}_t)+\sqrt{\beta_t} \mathbf{z}
\end{equation}
where $\alpha_t^{\prime}:=1 - \beta_t$ with $\beta_t$ as the given noise schedule for DDPM and  $\bar{\alpha}^{\prime}_t =  \prod_{s=1}^{t} \alpha^{\prime}_s$. 
Using the same techniques as above, we can trivially derive the corresponding iteration rules:
\begin{equation}
    \Var(\mathbf{x}_{t-1})=\frac{1}{\alpha^{\prime}_t}\Var(\mathbf{x}_t)-2\frac{1-\alpha^{\prime}_t}{\alpha_t^{\prime}\sqrt{1-\bar{\alpha}^{\prime}_t}}\Cov(\vx_t,\boldsymbol{\epsilon}_t)+\frac{(1-\alpha^{\prime}_t)^2}{\alpha^{\prime}_t(1-\bar{\alpha}^{\prime}_t)}\Var(\boldsymbol{\epsilon}_t)+\beta_t
\end{equation}
\begin{equation}
    \E(\vx_{t-1})=\frac{1}{\sqrt{\alpha_t^{\prime}}}\E(\vx_t)-\frac{1-\alpha_t^{\prime}}{\sqrt{\alpha^{\prime}_t(1-\bar{\alpha}^{\prime}_t)}}\E(\boldsymbol{\epsilon}_t)
\end{equation}
\cref{eq:mc} can still be leveraged to approximate $\Cov(\vx_t,\boldsymbol{\epsilon}_t)$. 
We iterate over these equations to obtain $\vx_0$ as well as its uncertainty $\Var(\vx_0)$. 
Since the advanced samplers, e.g., Analytic-DPM~\citep{bao2021analytic}, DDIM~\citep{song2020denoising} and DPM-Solver~\citep{lu2022dpm}, are more efficient and widely adopted, we also present the explicit formulations corresponding to them in Appendix~\ref{append:ddpm}. 
The 2-order DPM-solver is particularly popular in practice, e.g., in large-scale text-to-image models, but it is non-trivial to apply the above derivations directly to it because there is an extra hidden state introduced between $\vx_t$ and $\vx_{t-1}$. 
We propose to leverage structures like conditional independence to resolve this. 
Find more details in Appendix~\ref{append:ddpm}.

\begin{algorithm}[t]   
\vspace{-.1cm}
\caption{Pixel-wise uncertainty estimation via Bayesian inference. (BayesDiff)}
\label{alg:fulluq}
    \begin{algorithmic}[1]  
        \Require Starting point $\vx_T$, Monte Carlo sample size $S$, Pre-trained noise prediction model $\epsilon_\theta$.
        \Ensure  Image generation $\vx_0$ and pixel-wise uncertainty $\Var(\vx_0)$.
        \State Construct the pixel-wise variance prediction function ${\gamma}^2_\theta$ via LLLA;
            \State $\E(\vx_T)\gets\vx_T, \Var(\vx_T)\gets\mathbf{0}$, $\Cov(\vx_T, \boldsymbol{\epsilon}_T)\gets\mathbf{0}$;
        \For {$t=T \to 1$}
            \State Sample $\boldsymbol{\epsilon}_{t} \sim \mathcal{N}({\epsilon}_{\theta}(\vx_t, t), \mathrm{diag}({\gamma}^2_\theta(\vx_t, t)))$;
            \State Obtain $\vx_{t-1} $ via \cref{eq:disc}; %
            \State Estimate $\E(\vx_{t-1})$ and $\Var(\vx_{t-1})$ via \cref{eq:exp-sde} and \cref{eq:var-ite}; %
                \State sample $\vx_{t-1,i} \sim \mathcal{N}(\E(\vx_{t-1}), \Var(\vx_{t-1})), i=1,\dots, S$;
            \State Estimate $\Cov(\vx_{t-1},\boldsymbol{\epsilon}_{t-1}) $ via \cref{eq:mc}. %
        \EndFor
    \end{algorithmic}
\end{algorithm}

\vspace{-.1cm}
\textbf{Continuous-time reverse process.} 
Instead of quantifying the uncertainty captured by the discrete diffusion process as \cref{eq:disc}, we can also directly quantify that associated with the original continuous-time process, i.e., \cref{eq:sampling}.
We have derived an approximate expression illustrating the pattern of uncertainty dynamics at arbitrary timestep in $[0, T]$. However, estimating $\Var(\vx_0)$ using it is equally laborious as a discrete-time reverse process because discretization is still required to approximate the involved integration.
See Appendix~\ref{append:continuous} for more discussion.

\textbf{Algorithm.}
\cref{alg:fulluq} demonstrates the procedure of applying the developed uncertainty iteration principle to the SDE sampler in \cref{eq:disc}.
After obtaining the pixel-wise uncertainty $\Var(\vx_0)$, we can aggregate the elements into an image-wise metric for low-quality image filtering. 

\vspace{-.2cm}
\subsection{The Practical Acceleration}
\vspace{-.1cm}
\label{sec:fast}

\cref{alg:fulluq} computes the uncertainty of the hidden state at each sampling step, abbreviated as BayesDiff. 
The function ${\gamma}^2_\theta(\vx_t, t)$ produces outcomes along with $\epsilon_\theta(\vx_t, t)$, raising minimal added cost.
Yet, the MC estimation of $\Cov(\vx_t,\boldsymbol{\epsilon}_t) $ in \cref{eq:mc} causes $S$ (usually $S>10$) more evaluations of $\epsilon_\theta$, which can be prohibitive in the deployment scenarios.

To address this, we develop a faster variant of BayesDiff by performing uncertainty quantification on only a subset of the denoising steps rather than all of them, dubbed BayesDiff-Skip. 
Concretely, we pre-define a schedule $\tilde{\vt}:=\{\tilde{t}_1, \dots, \tilde{t}_U\}$ in advance.
For each timestep $t$, if $t \in \tilde{\vt}$, we sample $\boldsymbol{\epsilon}_{t}$ from LLLA and estimate corresponding uncertainty $\Var(\vx_{t-1})$ following the uncertainty iteration principle.
Otherwise, we adopt the deterministic sampling step where $\Cov(\vx_t, \boldsymbol{\epsilon}_{t})$ and $\Var(\boldsymbol{\epsilon}_{t})$ are set to zero.
We outline such a procedure in Appendix~\ref{append:bayesdiff}.

\begin{wrapfigure}{r}{0.4\textwidth}
\vspace{-.55cm}
  \begin{center}
\includegraphics[width=.36\textwidth]{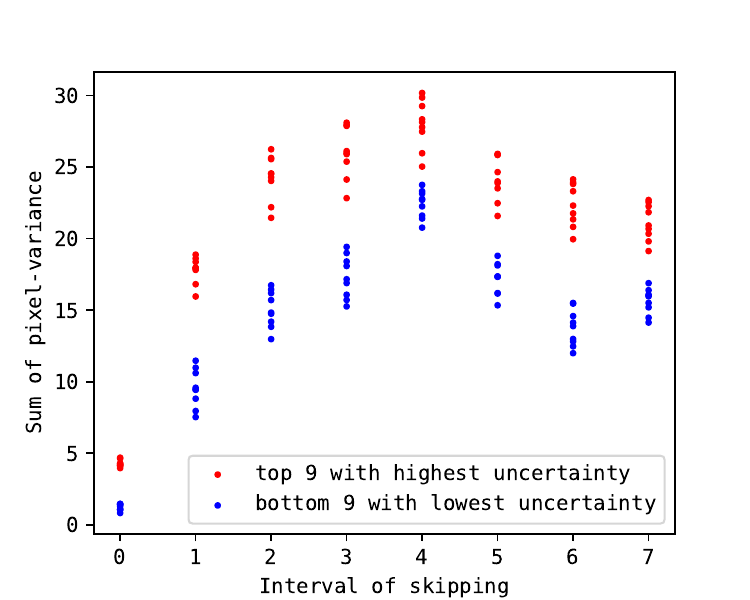}
  \end{center}
  \vspace{-.3cm}
  \caption{\small \footnotesize A study on the reliability of the BayesDiff-Skip algorithm. The top images with the highest uncertainty selected by BayesDiff are still with high uncertainty in BayesDiff-Skip algorithm.}
  \vspace{-.5cm}
  \label{fig:skip_step_change}
\end{wrapfigure}

\textbf{Consistency between BayesDiff-Skip and BayesDiff.} 
We check the reliability of BayesDiff-Skip here. 
Concretely, we generate 96 images using BayesDiff (skipping interval $=$ 0) under the DDIM sampling rule on ImageNet and mark the top 9 and bottom 9 samples with the highest and lowest uncertainty according to the summation of pixel-wise uncertainty. We send the same random noises (i.e., $\vx_T$) to BayesDiff-Skip (skipping interval $>$ 0) with various skipping intervals (unless specified otherwise, we evenly skip) to obtain generations close to the aforementioned 96 images yet with various uncertainties. 
We plot the uncertainties of the marked images in \cref{fig:skip_step_change}. 
It is shown that the marked images that BayesDiff is the most uncertain about remain the same for BayesDiff-Skip. Notably, in this experiment, BayesDiff-Skip can achieve a $5\times$ reduction in running time.

\vspace{-.1cm}
\section{Experiments}
\vspace{-.2cm}
In this section, we demonstrate the efficacy of BayesDiff in filtering out low-quality generations. Besides, 
we show that pixel-wise uncertainty can aid in generating diverse variations of the successful generation and addressing artifacts and misalignment of failure generations in text-to-image tasks. 
At last, we seek an intuitive understanding of the uncertainty estimates obtained by BayesDiff. %
We sum over the pixel-wise uncertainty to obtain an image-wise metric. 
Unless specified otherwise, we set the Monte Carlo sample size $S$ to $10$ and adopt BayesDiff-Skip with a skipping interval of 4, which makes our sampling and uncertainty quantification procedure consume no more than $2\times$ time than the vanilla sampling method.

\subsection{Effectiveness in Low-quality Image Filtering}

\label{sec:metric}
\begin{figure}[t]
\vspace{-.2cm}
\centering
\includegraphics[width=.95\linewidth]{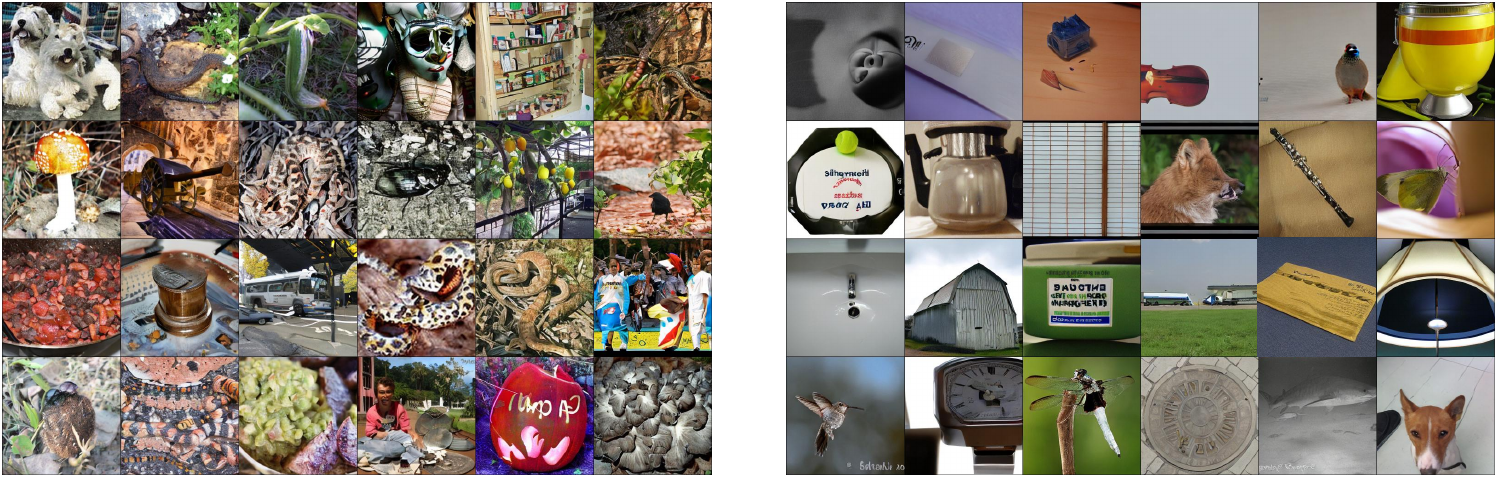}
\vspace{-.15cm}
\caption{The images with the highest (left) and lowest (right) uncertainty among 5000 unconditional generations of U-ViT model trained on ImageNet at $256\times 256$ resolution.}
\label{fig:imgnet_rank}
\end{figure}

\begin{figure}[t]
\centering
\includegraphics[width=.95\linewidth]{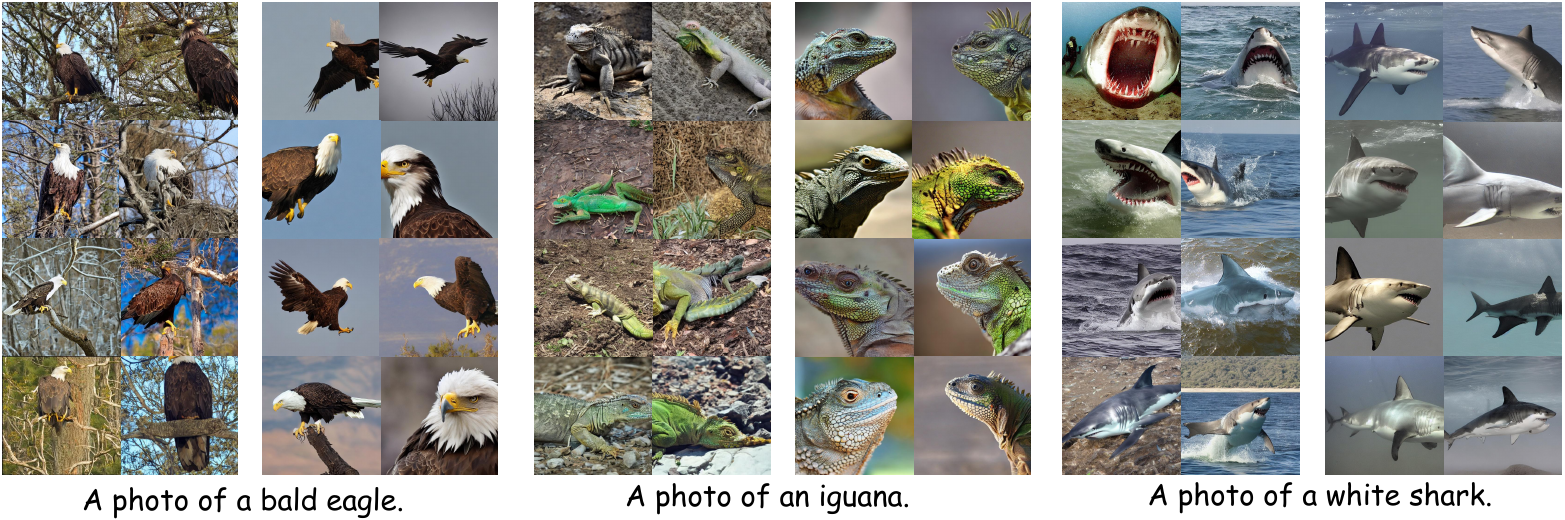}
\vspace{-.15cm}
\caption{The images with the highest (left) and lowest (right) uncertainty among 80 generations on Stable Diffusion at $512\times 512$ resolution.}
\label{fig:sd_rank}
\vspace{-.3cm}
\end{figure}

\textbf{Comparison between high and low uncertainty images.} 
We first conduct experiments on the U-ViT~\citep{bao2023all} model trained on ImageNet~\citep{deng2009imagenet} and Stable Diffusion, performing sampling and uncertainty quantification using BayesDiff-Skip. 
The sampling algorithm follows the 2-order DPM-Solver with 50 function evaluations (NFE).
We display the generations with the highest and lowest uncertainty in \cref{fig:imgnet_rank} and \cref{fig:sd_rank}. 


As shown above, our image-wise uncertainty metric is likely to indicate the level of clutter and the degree of subject prominence in the image. It can be used to detect low-quality images with cluttered backgrounds in downstream applications.

\begin{figure}[t]
\begin{center}
\includegraphics[width=1\textwidth]{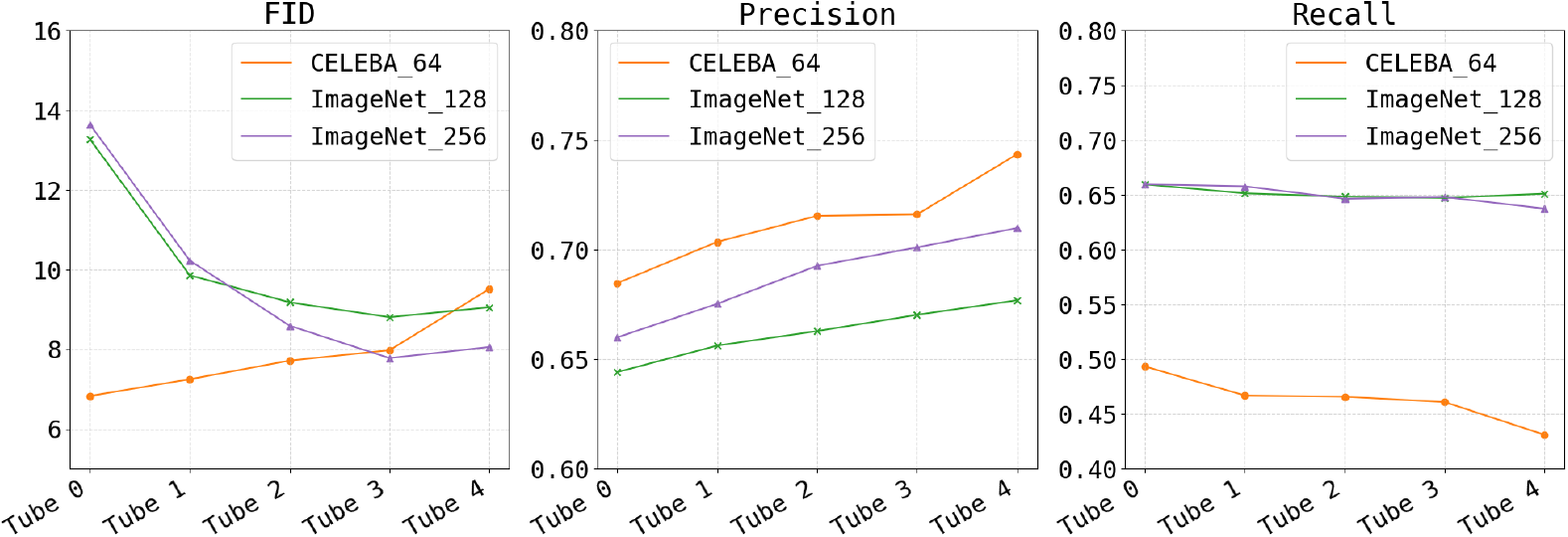}
\end{center}
\vspace{-8pt}
\caption{FID, Precision and Recall scores of 5 groups of generations with descending uncertainty on CELEBA and ImageNet datasets. Results show there is a strong correlation between our sample-wise uncertainty metric and traditional distributional metrics.}
\label{fig:tube}
\end{figure}

\textbf{Relationship between our sample-wise uncertainty metric and traditional distributional metrics.}
We use BayesDiff-Skip to generate 100,000 images on CELEBA~\citep{liu2015faceattributes} based on DDPM~\citep{ho2020denoising} model and DDIM sampler,  
250,000 $256\times 256$ ImageNet images based on U-ViT~\citep{bao2023all} and 2-order DPM-Solver, 
and 250,000 $128\times 128$ ImageNet images based on ADM~\citep{dhariwal2021diffusion} and DDIM. 
We separately divide the sets into five groups of the same size with descending uncertainty. 
We compute the traditional metrics for each group of data, including Precision~\citep{kynkaanniemi2019improved}, which evaluates the fidelity of the generation, Recall~\citep{kynkaanniemi2019improved}, which accounts for the diversity, and FID, which conjoins fidelity and diversity. 
We present the results in \cref{fig:tube}. 

\label{sec:5_tube_3_metric}

Notably,
images with higher uncertainty have higher Recalls in both scenarios, i.e., higher diversity, and images with lower uncertainty have higher Precisions, i.e., higher fidelity. 
This echoes the trade-off between Precision and Recall. 
Moreover, as \cref{fig:imgnet_rank} implies, images with high uncertainty have cluttered elements, which is consistent with the results of Precision. 
However, the trend of FID is different among the three datasets. 
The main reason is that the generations on the simple CELEBA are all good enough, so diversity becomes the main factor influencing FID. Conversely, on the more complex ImageNet, fidelity is the main factor influencing FID. 

\vspace{.5cm}
\begin{wrapfigure}{r}{0.42\textwidth}
\vspace{-0.45cm}
  \begin{center}
  \includegraphics[width=.42\textwidth]{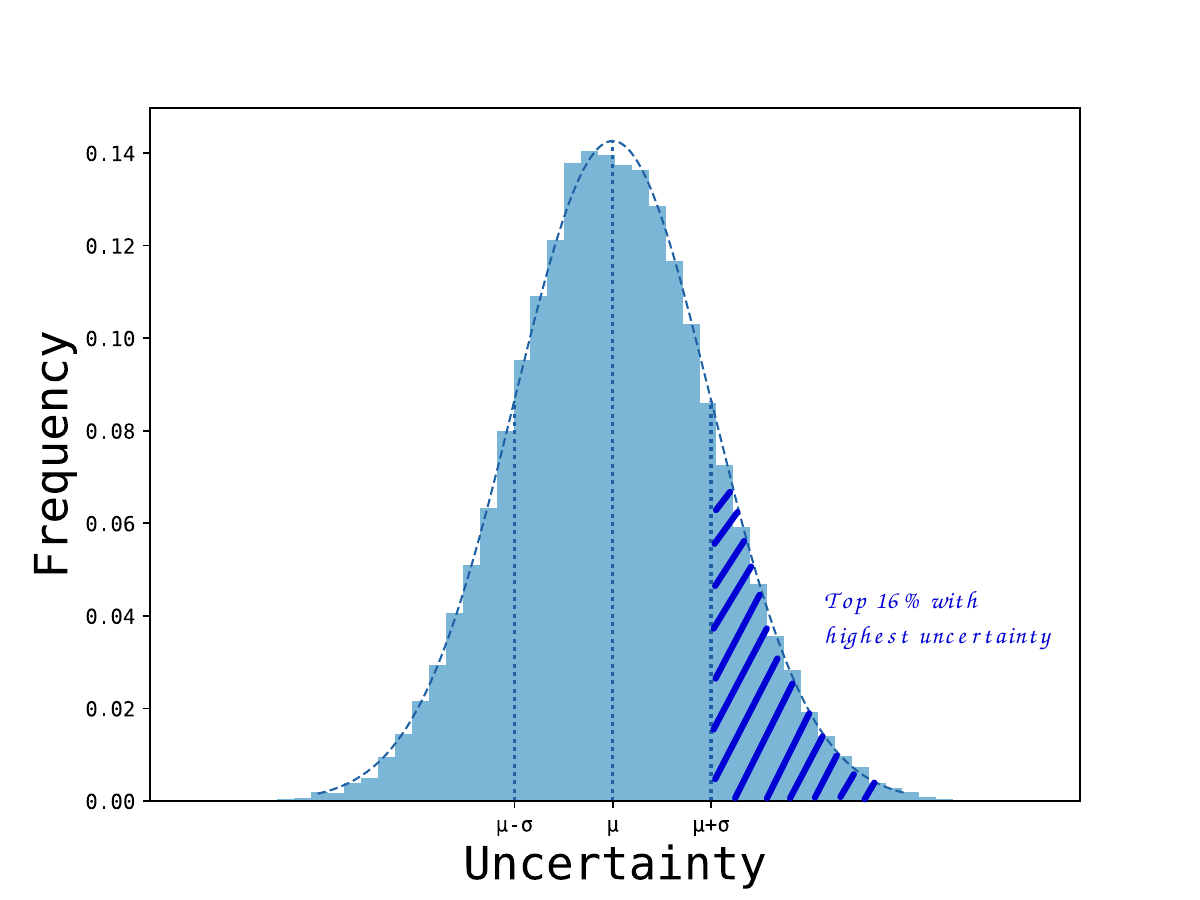}
  \end{center}
  \vspace{-0.2cm}
  \caption{The empirical distribution of the uncertainty estimates yielded by our approach. The dashed line denotes the normal distribution fitted on them. 
  Inspired by this, we propose to filter out the top $16\%$ samples with the highest uncertainty. }
  \label{fig:hist}
  \vspace{-.4cm}
\end{wrapfigure}
\textbf{Criterion for low-quality image filtering to improve the quality of generated distribution.}
We generate 50,000 images using a 2-order DPM-Solver sampler and BayesDiff-Skip on ImageNet at $256\times 256$ resolution to explore the distribution of image-wise uncertainty. As shown in \cref{fig:hist}, the empirical distribution of the image-wise uncertainty of the generations is approximately a normal distribution. According to the 3-sigma criterion of normal distribution for eliminating outliers, %
we eliminate low-quality images by filtering out the images with uncertainty higher than $\mu + \sigma$, which is equivalent to top $16\%$ images with the highest uncertainty. 
To test the effectiveness of this criterion, we generate 60,000 images with various models and samplers and select the 50,000 images with lower uncertainty. 
We compute the Precision, Recall, and FID of these samples in Table \ref{tab:metric}, which also includes the random selection baseline. 
The results validate that we filter out low-quality images precisely with such an uncertainty-based filtering criterion.

\begin{table}[t]
\vspace{-5pt}
\caption{Comparison on three metrics between randomly selected images and our selected images. We use 50 NFE for both DDIM and DPM-Solver sampler.}
\vspace{-.1cm}
\label{tab:metric}
\setlength{\tabcolsep}{0.55em} %
\small
\centering
\begin{tabular}{cccrrrrrr}
\toprule
\multirow{2}{*}{Model}&\multirow{2}{*}{Dataset}&\multirow{2}{*}{Sampler}& \multicolumn{2}{c}{\text{FID} $\downarrow$}& \multicolumn{2}{c}{\text{Precision} $\uparrow$} & \multicolumn{2}{c}{\text{Recall} $\uparrow$} \\
\cmidrule{4-9} &&& random & ours & random & ours & random & ours \\
\midrule
ADM&ImageNet 128&DDIM& $8.68 \pm 0.04$ &8.48 &0.661 &0.665 &0.655 &0.653 \\
ADM&ImageNet 128&2-order DPM-Solver& $9.77 \pm 0.03$ &9.67 & 0.657 &0.659 &0.649 &0.649\\
\hline
U-ViT&ImageNet 256&2-order DPM-Solver& $7.24 \pm 0.02$& 6.81 & 0.698&0.705& 0.658&0.657 \\
U-ViT&ImageNet 512&2-order DPM-Solver& $17.72 \pm 0.03$& 16.87 &0.728&0.732& 0.602&0.604\\
\bottomrule
\end{tabular}
\vspace{-.3cm}
\end{table}

\vspace{-.5cm}
\subsection{Pixel-wise Uncertainty: A Tool for Diversity Enhancement and Artifact Rectification}
\begin{figure}[t]
\vspace{-.9cm}
\begin{center}
\includegraphics[width=.95\textwidth]{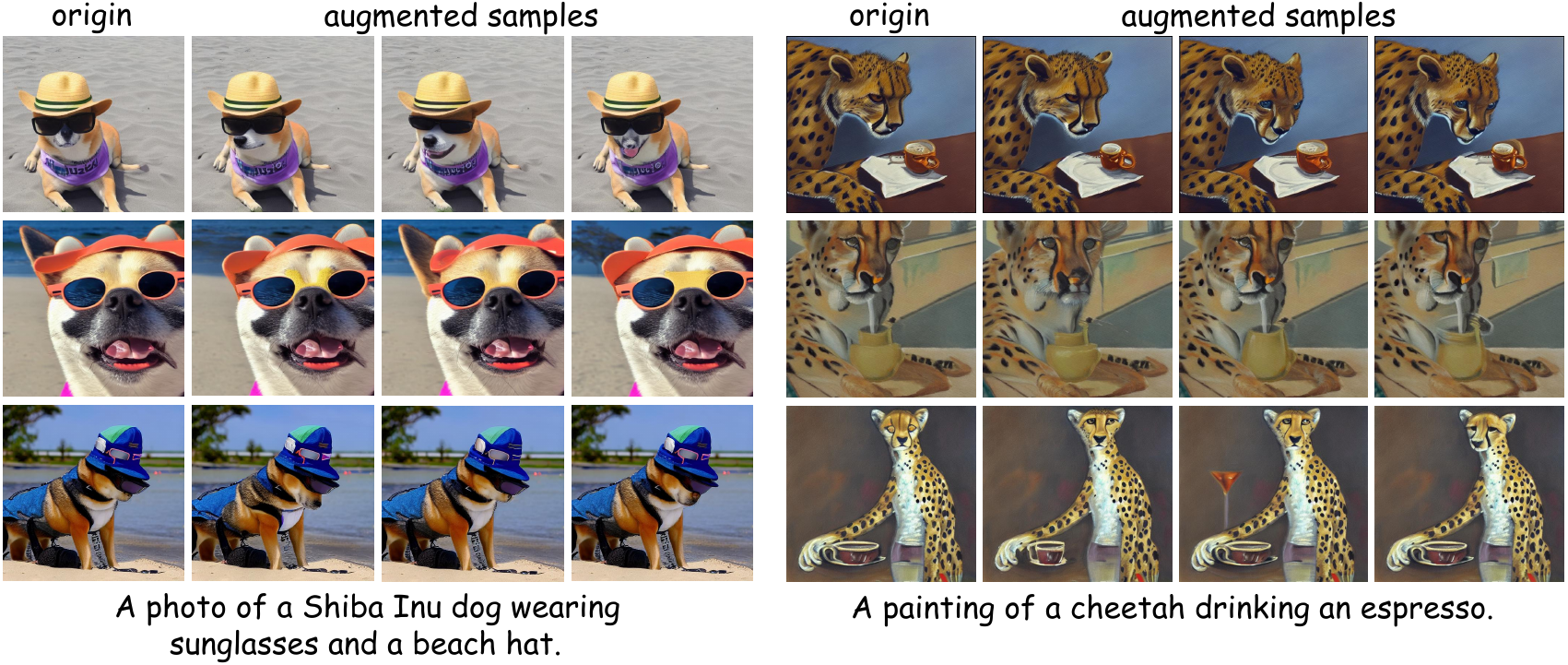}
\end{center}
\vspace{-.2cm}
\caption{Examples of the augmentation of good generations with enhanced diversity on Stable Diffusion with DDIM sampler (50 NFE).}
\label{fig:aug}
\end{figure}

\begin{figure}[t]
\begin{center}
\includegraphics[width=.95\textwidth]{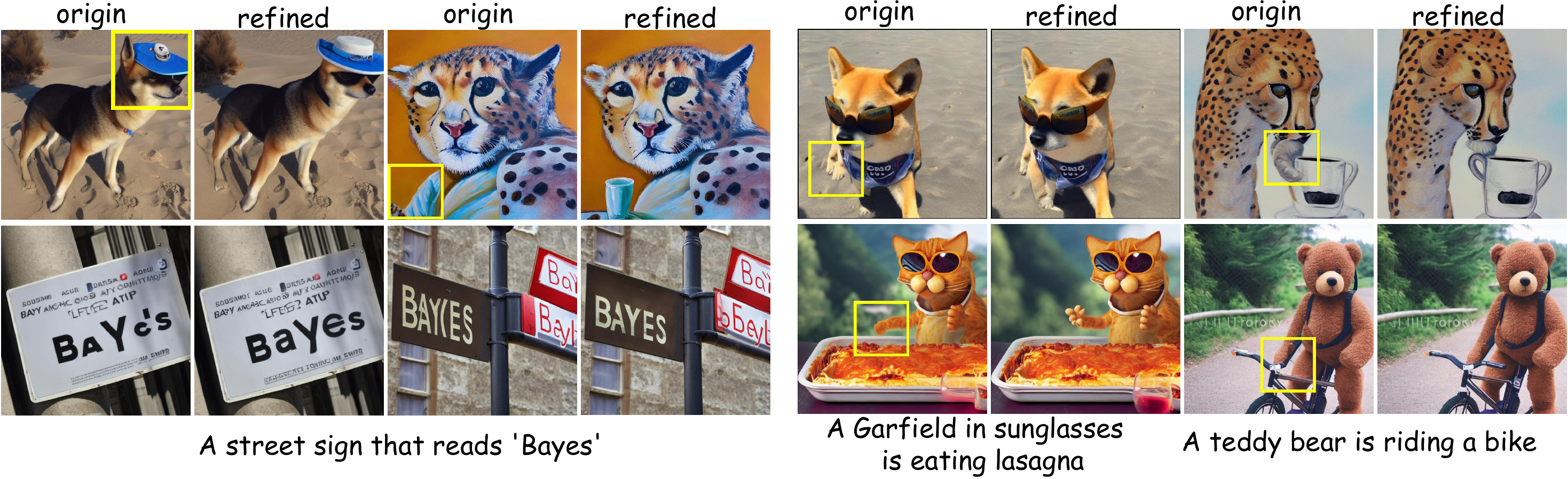}
\end{center}
\vspace{-.2cm}
\caption{Examples of the rectification of artifacts and misalignment in failure generations on Stable Diffusion with DDIM sampler (50 NFE). The flawed samples are identified by humans and the bounding boxes are manually annotated.}
\label{fig:correctness}
\end{figure}

The main issue with Stable Diffusion is that usually, only a few generations denoised from `good' initial noises can align with the input textual description~\citep{wu2023better}. 
We show that the pixel-wise uncertainty estimated by BayesDiff can be leveraged to alleviate such an issue. 
Specifically, BayesDiff-Skip allows for introducing a distribution of $\vx_t \sim \mathcal{N}(\E(\vx_{t}), \Var(\vx_t))$ for any time $t \in [0, T]$. 
If the final sample $\vx_0$ appears to be a good generation, we can resample $\vx_{t,i}$ from $\mathcal{N}(\E(\vx_{t}), \Var(\vx_t))$ and denoise it to a new one $\vx_{0,i}$, which is similar yet different from $\vx_0$. 
We leverage BayesDiff-Skip with a skipping interval of 1 and DDIM sampler with 50 NFE to test this.
We use the Gaussian distribution estimated at $t=40$ for resampling. \cref{fig:aug} shows that the resampled $\vx_{0,i}$ still conforms the above hypothesis. 
Moreover, \cref{fig:correctness} shows that in some cases, the artifacts in original samples can be rectified, and hence the failed samples, which are mismatched with the prompts, are rectified into successful samples. More examples are shown in Appendix~\ref{append:more_examples}.

\vspace{-.1cm}
\label{sec:bootstrap}

\subsection{Further Understanding of Pixel-wise Uncertainty}
\label{sec:vis}
\begin{figure}[t]
\vspace{-.6cm}
\begin{center}
\includegraphics[width=0.85\textwidth]{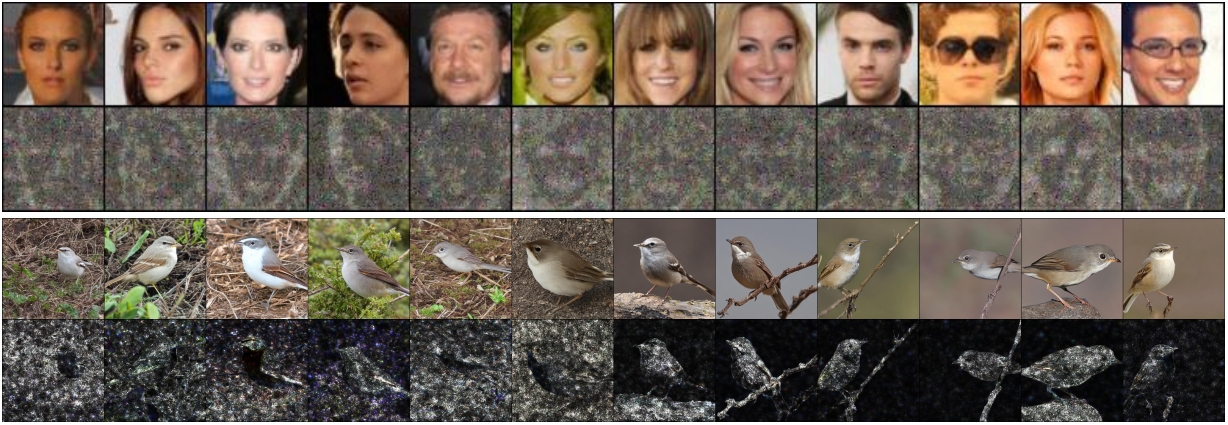}
\end{center}
\vspace{-.3cm}
\caption{Visualization of the pixel-wise uncertainty of generations on CELEBA (top) and from Stable Diffusion with prompt `A photo of a whitethroat' (bottom). We adopt BayesDiff-Skip and DDIM sampler (50 NFE) in both cases.}
\label{fig:vis_uq}
\end{figure}

\textbf{Visualization of pixel-wise uncertainty.} To gain an intuitive understanding of the obtained uncertainty estimates, we visualize them in this section. %
In detail, we launch BayesDiff using DDPM model for generating CELEBA images
and Stable Diffusion for generating prompt-conditional images. 
Nonetheless, the visualization for Stable Diffusion is not straightforward because we actually obtain the variance in the final latent states. 
To solve this problem, we sample a variety of latent states and send them to the decoder of Stable Diffusion.
We estimate the empirical variance of the outcomes as the final pixel-wise uncertainty. 
\cref{fig:vis_uq} presents the results. As shown, our uncertainty estimates carry semantic information. 
The eyes, noses, and mouths of human faces demonstrate greater uncertainty on CELEBA,
and the contours of objects in images from Stable Diffusion exhibit higher levels of uncertainty. 
This further explains the higher image-wise uncertainty for clutter images and the lower one for clean images mentioned in \Secref{sec:metric}.

\begin{figure}[t]
\begin{center}
\includegraphics[width=0.85\textwidth]{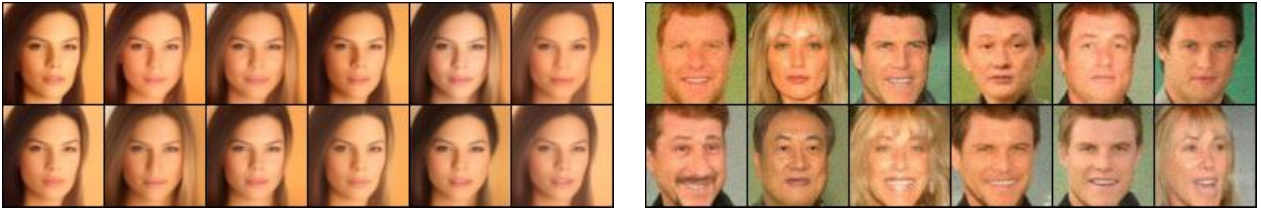}
\end{center}
\vspace{-.3cm}
\caption{Visualization of $\vx_{adjacent, 0}$, which are denoised from adjacent initial positions corresponding to low image-wise uncertainty (left, ranges from 2.5 to 2.8) and high one (right, ranges from 24.5 to 28.1) on CELEBA with DDIM sampler (50 NFE). }
\vspace{-.2cm}
\label{fig:adjacent}
\end{figure}

\vspace{-.1cm}
\textbf{Visualization of generations of adjacent initial positions.} In BayesDiff, the estimated uncertainty $\Var(\vx_0)$ is dependent on the corresponding initial position $\vx_T \sim \mathcal{N}(\mathbf{0}, \mathbf{I})$,
and higher uncertainty estimates correspond to larger variations in the sampling trajectory. 
Therefore, adding minor perturbation to initial positions corresponding to high uncertainty should produce diverse generations. 
We conduct experiments on CELEBA~\citep{liu2015faceattributes} with DDIM sampler to validate this conjecture. 
Specifically, we define the adjacent initial position $\vx_{adjacent, T} := \sqrt{1-\eta} \vx_T 
 + \sqrt{\eta} \vz, \vz \sim \mathcal{N}(0,I)$ and visualize $\vx_{adjacent, 0}$, which is denoised from $\vx_{adjacent, T}$. 
\cref{fig:adjacent} shows that images generated from adjacent positions with high uncertainty indeed exhibit greater diversity. 
This result also echos the higher Recall of the group with higher uncertainty on CELEBA in \cref{sec:5_tube_3_metric}.
\vspace{-.2cm}

\section{Related Work}
\vspace{-.15cm}
Several works incorporate Baysian inference into deep generative models and exihibit strong performance.
Variational Autoencoder (VAE)~\citep{kingma2014auto} is a classic generative model that learns the data distribution through Bayesian variational inference. VAEs have been applied in various domains and demonstrate powerful capabilities of data representation and generation.~\citep{kingma2014auto, hou2017deep, bowman2015generating,semeniuta2017hybrid,wang2019topic, ha2018world} 
Variational Diffusion Models (VDMs)~\citep{kingma2021variational} employ a signal-to-noise ratio function to parameterize the forward noise schedule in diffusion models, enabling the direct optimization of the variational lower bound (VLB) and accelerating training a lot.

\section{Conclusion}
\vspace{-.15cm}
In this paper, we introduce BayesDiff, a framework for pixel-wise uncertainty estimation in Diffusion Models via Bayesian inference. 
We have empirically demonstrated that the estimated pixel-wise uncertainty holds a significant practical value, including being utilized as an image-wise uncertainty metric for filtering low-quality images and a tool for diversity enhancement and misalignment rectification in text-to-image generations. 
Apart from image generations, the powerful capability of Diffusion Models to generate realistic samples has been applied in various other domains such as natural language processing (audio synthesis~\citep{huang2022prodiff}; text-to-speech~\citep{kim2022guided}) and AI for science (molecular conformation prediction~\citep{xu2021geodiff}; material design~\citep{luo2022antigen}). 
We believe BayesDiff holds great potential for incorporating with these applications to improve the predictive uncertainty and calibrate generations.

\section*{Ethics Statement}
This work is a fundamental research in machine learning, the potential negative consequences are
not apparent. While it is theoretically possible for any technique to be misused, the likelihood of
such misuse occurring at the current stage is low.

\section*{Acknowledgments}
This work was supported by NSF of China (Nos. 62306176, 62076145), Natural Science Foundation of Shanghai (No. 23ZR1428700), the
Key Research and Development Program of Shandong Province, China (No. 2023CXGC010112), and Beijing Outstanding Young Scientist Program (No. BJJWZYJH012019100020098). C. Li was also sponsored by Beijing Nova Program (No. 20220484044).

\bibliography{iclr2024_conference}

\begin{thebibliography}{47}
\providecommand{\natexlab}[1]{#1}
\providecommand{\url}[1]{\texttt{#1}}
\expandafter\ifx\csname urlstyle\endcsname\relax
  \providecommand{\doi}[1]{doi: #1}\else
  \providecommand{\doi}{doi: \begingroup \urlstyle{rm}\Url}\fi

\bibitem[Bao et~al.(2021)Bao, Li, Zhu, and Zhang]{bao2021analytic}
Fan Bao, Chongxuan Li, Jun Zhu, and Bo~Zhang.
\newblock Analytic-dpm: an analytic estimate of the optimal reverse variance in
  diffusion probabilistic models.
\newblock In \emph{International Conference on Learning Representations}, 2021.

\bibitem[Bao et~al.(2023)Bao, Nie, Xue, Cao, Li, Su, and Zhu]{bao2023all}
Fan Bao, Shen Nie, Kaiwen Xue, Yue Cao, Chongxuan Li, Hang Su, and Jun Zhu.
\newblock All are worth words: A vit backbone for diffusion models.
\newblock In \emph{Proceedings of the IEEE/CVF Conference on Computer Vision
  and Pattern Recognition}, pp.\  22669--22679, 2023.

\bibitem[Blundell et~al.(2015)Blundell, Cornebise, Kavukcuoglu, and
  Wierstra]{blundell2015weight}
Charles Blundell, Julien Cornebise, Koray Kavukcuoglu, and Daan Wierstra.
\newblock Weight uncertainty in neural network.
\newblock In \emph{International Conference on Machine Learning}, pp.\
  1613--1622, 2015.

\bibitem[Bowman et~al.(2015)Bowman, Vilnis, Vinyals, Dai, Jozefowicz, and
  Bengio]{bowman2015generating}
Samuel~R Bowman, Luke Vilnis, Oriol Vinyals, Andrew~M Dai, Rafal Jozefowicz,
  and Samy Bengio.
\newblock Generating sentences from a continuous space.
\newblock \emph{arXiv preprint arXiv:1511.06349}, 2015.

\bibitem[Chen et~al.(2014)Chen, Fox, and Guestrin]{chen2014stochastic}
Tianqi Chen, Emily Fox, and Carlos Guestrin.
\newblock Stochastic gradient hamiltonian monte carlo.
\newblock In \emph{International Conference on Machine Learning}, pp.\
  1683--1691. PMLR, 2014.

\bibitem[Daxberger et~al.(2021{\natexlab{a}})Daxberger, Kristiadi, Immer,
  Eschenhagen, Bauer, and Hennig]{daxberger2021laplace}
Erik Daxberger, Agustinus Kristiadi, Alexander Immer, Runa Eschenhagen,
  Matthias Bauer, and Philipp Hennig.
\newblock Laplace redux-effortless bayesian deep learning.
\newblock \emph{Advances in Neural Information Processing Systems},
  34:\penalty0 20089--20103, 2021{\natexlab{a}}.

\bibitem[Daxberger et~al.(2021{\natexlab{b}})Daxberger, Nalisnick, Allingham,
  Antor{\'a}n, and Hern{\'a}ndez-Lobato]{daxberger2021bayesian}
Erik Daxberger, Eric Nalisnick, James~U Allingham, Javier Antor{\'a}n, and
  Jos{\'e}~Miguel Hern{\'a}ndez-Lobato.
\newblock Bayesian deep learning via subnetwork inference.
\newblock In \emph{International Conference on Machine Learning}, pp.\
  2510--2521. PMLR, 2021{\natexlab{b}}.

\bibitem[Deng et~al.(2009)Deng, Dong, Socher, Li, Li, and
  Fei-Fei]{deng2009imagenet}
Jia Deng, Wei Dong, Richard Socher, Li-Jia Li, Kai Li, and Li~Fei-Fei.
\newblock Imagenet: A large-scale hierarchical image database.
\newblock In \emph{2009 IEEE conference on computer vision and pattern
  recognition}, pp.\  248--255. Ieee, 2009.

\bibitem[Deng et~al.(2021)Deng, Yang, Xu, Su, and Zhu]{deng2021libre}
Zhijie Deng, Xiao Yang, Shizhen Xu, Hang Su, and Jun Zhu.
\newblock Libre: A practical bayesian approach to adversarial detection.
\newblock In \emph{Proceedings of the IEEE/CVF conference on computer vision
  and pattern recognition}, pp.\  972--982, 2021.

\bibitem[Dhariwal \& Nichol(2021)Dhariwal and Nichol]{dhariwal2021diffusion}
Prafulla Dhariwal and Alexander Nichol.
\newblock Diffusion models beat gans on image synthesis.
\newblock \emph{Advances in neural information processing systems},
  34:\penalty0 8780--8794, 2021.

\bibitem[Foong et~al.(2019)Foong, Li, Hern{\'a}ndez-Lobato, and
  Turner]{foong2019between}
Andrew~YK Foong, Yingzhen Li, Jos{\'e}~Miguel Hern{\'a}ndez-Lobato, and
  Richard~E Turner.
\newblock 'in-between'uncertainty in bayesian neural networks.
\newblock \emph{arXiv preprint arXiv:1906.11537}, 2019.

\bibitem[Gu et~al.(2022)Gu, Chen, Bao, Wen, Zhang, Chen, Yuan, and
  Guo]{gu2022vector}
Shuyang Gu, Dong Chen, Jianmin Bao, Fang Wen, Bo~Zhang, Dongdong Chen, Lu~Yuan,
  and Baining Guo.
\newblock Vector quantized diffusion model for text-to-image synthesis.
\newblock In \emph{Proceedings of the IEEE/CVF Conference on Computer Vision
  and Pattern Recognition}, pp.\  10696--10706, 2022.

\bibitem[Ha \& Schmidhuber(2018)Ha and Schmidhuber]{ha2018world}
David Ha and J{\"u}rgen Schmidhuber.
\newblock World models.
\newblock \emph{arXiv preprint arXiv:1803.10122}, 2018.

\bibitem[Hern{\'a}ndez-Lobato \& Adams(2015)Hern{\'a}ndez-Lobato and
  Adams]{hernandez2015probabilistic}
Jos{\'e}~Miguel Hern{\'a}ndez-Lobato and Ryan Adams.
\newblock Probabilistic backpropagation for scalable learning of {B}ayesian
  neural networks.
\newblock In \emph{International Conference on Machine Learning}, pp.\
  1861--1869, 2015.

\bibitem[Heusel et~al.(2017)Heusel, Ramsauer, Unterthiner, Nessler, and
  Hochreiter]{heusel2017gans}
Martin Heusel, Hubert Ramsauer, Thomas Unterthiner, Bernhard Nessler, and Sepp
  Hochreiter.
\newblock Gans trained by a two time-scale update rule converge to a local nash
  equilibrium.
\newblock \emph{Advances in neural information processing systems}, 30, 2017.

\bibitem[Ho et~al.(2020)Ho, Jain, and Abbeel]{ho2020denoising}
Jonathan Ho, Ajay Jain, and Pieter Abbeel.
\newblock Denoising diffusion probabilistic models.
\newblock \emph{Advances in Neural Information Processing Systems},
  33:\penalty0 6840--6851, 2020.

\bibitem[Hou et~al.(2017)Hou, Shen, Sun, and Qiu]{hou2017deep}
Xianxu Hou, Linlin Shen, Ke~Sun, and Guoping Qiu.
\newblock Deep feature consistent variational autoencoder.
\newblock In \emph{2017 IEEE winter conference on applications of computer
  vision (WACV)}, pp.\  1133--1141. IEEE, 2017.

\bibitem[Huang et~al.(2022)Huang, Zhao, Liu, Liu, Cui, and
  Ren]{huang2022prodiff}
Rongjie Huang, Zhou Zhao, Huadai Liu, Jinglin Liu, Chenye Cui, and Yi~Ren.
\newblock Prodiff: Progressive fast diffusion model for high-quality
  text-to-speech.
\newblock In \emph{Proceedings of the 30th ACM International Conference on
  Multimedia}, pp.\  2595--2605, 2022.

\bibitem[Karras et~al.(2022)Karras, Aittala, Aila, and
  Laine]{karras2022elucidating}
Tero Karras, Miika Aittala, Timo Aila, and Samuli Laine.
\newblock Elucidating the design space of diffusion-based generative models.
\newblock \emph{Advances in Neural Information Processing Systems},
  35:\penalty0 26565--26577, 2022.

\bibitem[Khan et~al.(2018)Khan, Nielsen, Tangkaratt, Lin, Gal, and
  Srivastava]{khan2018fast}
Mohammad~Emtiyaz Khan, Didrik Nielsen, Voot Tangkaratt, Wu~Lin, Yarin Gal, and
  Akash Srivastava.
\newblock Fast and scalable {B}ayesian deep learning by weight-perturbation in
  adam.
\newblock In \emph{International Conference on Machine Learning}, pp.\
  2616--2625, 2018.

\bibitem[Kim et~al.(2022)Kim, Kim, and Yoon]{kim2022guided}
Heeseung Kim, Sungwon Kim, and Sungroh Yoon.
\newblock Guided-tts: A diffusion model for text-to-speech via classifier
  guidance.
\newblock In \emph{International Conference on Machine Learning}, pp.\
  11119--11133. PMLR, 2022.

\bibitem[Kingma et~al.(2021)Kingma, Salimans, Poole, and
  Ho]{kingma2021variational}
Diederik Kingma, Tim Salimans, Ben Poole, and Jonathan Ho.
\newblock Variational diffusion models.
\newblock \emph{Advances in neural information processing systems},
  34:\penalty0 21696--21707, 2021.

\bibitem[Kingma \& Welling(2014)Kingma and Welling]{kingma2014auto}
Diederik~P Kingma and Max Welling.
\newblock Auto-encoding variational bayes.
\newblock \emph{stat}, 1050:\penalty0 1, 2014.

\bibitem[Kristiadi et~al.(2020)Kristiadi, Hein, and Hennig]{kristiadi2020being}
Agustinus Kristiadi, Matthias Hein, and Philipp Hennig.
\newblock Being bayesian, even just a bit, fixes overconfidence in relu
  networks.
\newblock In \emph{International conference on machine learning}, pp.\
  5436--5446. PMLR, 2020.

\bibitem[Kynk{\"a}{\"a}nniemi et~al.(2019)Kynk{\"a}{\"a}nniemi, Karras, Laine,
  Lehtinen, and Aila]{kynkaanniemi2019improved}
Tuomas Kynk{\"a}{\"a}nniemi, Tero Karras, Samuli Laine, Jaakko Lehtinen, and
  Timo Aila.
\newblock Improved precision and recall metric for assessing generative models.
\newblock \emph{Advances in Neural Information Processing Systems}, 32, 2019.

\bibitem[Liu \& Wang(2016)Liu and Wang]{liu2016stein}
Qiang Liu and Dilin Wang.
\newblock Stein variational gradient descent: A general purpose {B}ayesian
  inference algorithm.
\newblock In \emph{Advances in Neural Information Processing Systems}, pp.\
  2378--2386, 2016.

\bibitem[Liu et~al.(2015)Liu, Luo, Wang, and Tang]{liu2015faceattributes}
Ziwei Liu, Ping Luo, Xiaogang Wang, and Xiaoou Tang.
\newblock Deep learning face attributes in the wild.
\newblock In \emph{Proceedings of International Conference on Computer Vision
  (ICCV)}, December 2015.

\bibitem[Louizos \& Welling(2016)Louizos and Welling]{louizos2016structured}
Christos Louizos and Max Welling.
\newblock Structured and efficient variational deep learning with matrix
  gaussian posteriors.
\newblock In \emph{International Conference on Machine Learning}, pp.\
  1708--1716, 2016.

\bibitem[Lu et~al.(2022)Lu, Zhou, Bao, Chen, Li, and Zhu]{lu2022dpm}
Cheng Lu, Yuhao Zhou, Fan Bao, Jianfei Chen, Chongxuan Li, and Jun Zhu.
\newblock Dpm-solver: A fast ode solver for diffusion probabilistic model
  sampling in around 10 steps.
\newblock \emph{Advances in Neural Information Processing Systems},
  35:\penalty0 5775--5787, 2022.

\bibitem[Lugmayr et~al.(2022)Lugmayr, Danelljan, Romero, Yu, Timofte, and
  Van~Gool]{lugmayr2022repaint}
Andreas Lugmayr, Martin Danelljan, Andres Romero, Fisher Yu, Radu Timofte, and
  Luc Van~Gool.
\newblock Repaint: Inpainting using denoising diffusion probabilistic models.
\newblock In \emph{Proceedings of the IEEE/CVF Conference on Computer Vision
  and Pattern Recognition}, pp.\  11461--11471, 2022.

\bibitem[Luo et~al.(2022)Luo, Su, Peng, Wang, Peng, and Ma]{luo2022antigen}
Shitong Luo, Yufeng Su, Xingang Peng, Sheng Wang, Jian Peng, and Jianzhu Ma.
\newblock Antigen-specific antibody design and optimization with
  diffusion-based generative models for protein structures.
\newblock \emph{Advances in Neural Information Processing Systems},
  35:\penalty0 9754--9767, 2022.

\bibitem[Mackay(1992)]{mackay1992bayesian}
David John~Cameron Mackay.
\newblock \emph{Bayesian methods for adaptive models}.
\newblock PhD thesis, California Institute of Technology, 1992.

\bibitem[Maddox et~al.(2019)Maddox, Izmailov, Garipov, Vetrov, and
  Wilson]{maddox2019simple}
Wesley~J Maddox, Pavel Izmailov, Timur Garipov, Dmitry~P Vetrov, and
  Andrew~Gordon Wilson.
\newblock A simple baseline for bayesian uncertainty in deep learning.
\newblock \emph{Advances in neural information processing systems}, 32, 2019.

\bibitem[Ritter et~al.(2018)Ritter, Botev, and Barber]{ritter2018scalable}
Hippolyt Ritter, Aleksandar Botev, and David Barber.
\newblock A scalable laplace approximation for neural networks.
\newblock In \emph{6th International Conference on Learning Representations},
  volume~6. International Conference on Representation Learning, 2018.

\bibitem[Rombach et~al.(2022)Rombach, Blattmann, Lorenz, Esser, and
  Ommer]{rombach2022high}
Robin Rombach, Andreas Blattmann, Dominik Lorenz, Patrick Esser, and Bj{\"o}rn
  Ommer.
\newblock High-resolution image synthesis with latent diffusion models.
\newblock In \emph{Proceedings of the IEEE/CVF conference on computer vision
  and pattern recognition}, pp.\  10684--10695, 2022.

\bibitem[Saharia et~al.(2022)Saharia, Chan, Saxena, Li, Whang, Denton,
  Ghasemipour, Gontijo~Lopes, Karagol~Ayan, Salimans,
  et~al.]{saharia2022photorealistic}
Chitwan Saharia, William Chan, Saurabh Saxena, Lala Li, Jay Whang, Emily~L
  Denton, Kamyar Ghasemipour, Raphael Gontijo~Lopes, Burcu Karagol~Ayan, Tim
  Salimans, et~al.
\newblock Photorealistic text-to-image diffusion models with deep language
  understanding.
\newblock \emph{Advances in Neural Information Processing Systems},
  35:\penalty0 36479--36494, 2022.

\bibitem[Salimans et~al.(2016)Salimans, Goodfellow, Zaremba, Cheung, Radford,
  and Chen]{salimans2016improved}
Tim Salimans, Ian Goodfellow, Wojciech Zaremba, Vicki Cheung, Alec Radford, and
  Xi~Chen.
\newblock Improved techniques for training gans.
\newblock \emph{Advances in neural information processing systems}, 29, 2016.

\bibitem[Semeniuta et~al.(2017)Semeniuta, Severyn, and
  Barth]{semeniuta2017hybrid}
Stanislau Semeniuta, Aliaksei Severyn, and Erhardt Barth.
\newblock A hybrid convolutional variational autoencoder for text generation.
\newblock \emph{arXiv preprint arXiv:1702.02390}, 2017.

\bibitem[Song et~al.(2020)Song, Meng, and Ermon]{song2020denoising}
Jiaming Song, Chenlin Meng, and Stefano Ermon.
\newblock Denoising diffusion implicit models.
\newblock In \emph{International Conference on Learning Representations}, 2020.

\bibitem[Song et~al.(2021)Song, Sohl-Dickstein, Kingma, Kumar, Ermon, and
  Poole]{songscore}
Yang Song, Jascha Sohl-Dickstein, Diederik~P Kingma, Abhishek Kumar, Stefano
  Ermon, and Ben Poole.
\newblock Score-based generative modeling through stochastic differential
  equations.
\newblock In \emph{International Conference on Learning Representations}, 2021.

\bibitem[Wang et~al.(2019)Wang, Gan, Xu, Zhang, Wang, Shen, Chen, and
  Carin]{wang2019topic}
Wenlin Wang, Zhe Gan, Hongteng Xu, Ruiyi Zhang, Guoyin Wang, Dinghan Shen,
  Changyou Chen, and Lawrence Carin.
\newblock Topic-guided variational autoencoders for text generation.
\newblock In \emph{NAACL HLT 2019-2019 Conference of the North American Chapter
  of the Association for Computational Linguistics: Human Language
  Technologies-Proceedings of the Conference}, pp.\  166--177. Association for
  Computational Linguistics (ACL), 2019.

\bibitem[Welling \& Teh(2011)Welling and Teh]{welling2011bayesian}
Max Welling and Yee~W Teh.
\newblock Bayesian learning via stochastic gradient langevin dynamics.
\newblock In \emph{International Conference on Machine Learning}, pp.\
  681--688, 2011.

\bibitem[Wu et~al.(2023)Wu, Sun, Zhu, Zhao, and Li]{wu2023better}
Xiaoshi Wu, Keqiang Sun, Feng Zhu, Rui Zhao, and Hongsheng Li.
\newblock Better aligning text-to-image models with human preference.
\newblock \emph{arXiv e-prints}, pp.\  arXiv--2303, 2023.

\bibitem[Xu et~al.(2021)Xu, Yu, Song, Shi, Ermon, and Tang]{xu2021geodiff}
Minkai Xu, Lantao Yu, Yang Song, Chence Shi, Stefano Ermon, and Jian Tang.
\newblock Geodiff: A geometric diffusion model for molecular conformation
  generation.
\newblock In \emph{International Conference on Learning Representations}, 2021.

\bibitem[Zhang et~al.(2023)Zhang, Zhang, Zhang, and Kweon]{zhang2023text}
Chenshuang Zhang, Chaoning Zhang, Mengchun Zhang, and In~So Kweon.
\newblock Text-to-image diffusion model in generative ai: A survey.
\newblock \emph{arXiv preprint arXiv:2303.07909}, 2023.

\bibitem[Zhang et~al.(2018)Zhang, Sun, Duvenaud, and Grosse]{zhang2018noisy}
Guodong Zhang, Shengyang Sun, David Duvenaud, and Roger Grosse.
\newblock Noisy natural gradient as variational inference.
\newblock In \emph{International Conference on Machine Learning}, pp.\
  5847--5856, 2018.

\bibitem[Zhang et~al.(2019)Zhang, Li, Zhang, Chen, and
  Wilson]{zhang2019cyclical}
Ruqi Zhang, Chunyuan Li, Jianyi Zhang, Changyou Chen, and Andrew~Gordon Wilson.
\newblock Cyclical stochastic gradient mcmc for bayesian deep learning.
\newblock \emph{arXiv preprint arXiv:1902.03932}, 2019.

\end{thebibliography}
\bibliographystyle{iclr2024_conference}
\newpage
\appendix
\section{Appendix}
\subsection{Iteration Rules for Other Sampling Methods}
\label{append:ddpm}
For Analytic-DPM \citep{bao2021analytic}, the sampling method is: \begin{equation}
\mathbf{x}_{t-1}=\frac{1}{\sqrt{\alpha^{\prime}_t}}(\mathbf{x}_t-\frac{1-\alpha^{\prime}_t}{\sqrt{1-\bar{\alpha}^{\prime}_t}}\boldsymbol{\epsilon}_t)+\sigma_t \mathbf{z}
\end{equation}
\begin{equation}
\sigma_t^2 = \lambda_t^2 +(\sqrt{\frac{1-\bar{\alpha}^{\prime}_t}{\alpha^{\prime}_t}}-\sqrt{1-\bar{\alpha}_{t-1}^{\prime}-\lambda_t^2})^2(1-(1-\bar{\alpha}_t^{\prime})\Gamma_t)
\end{equation} 
\begin{equation}
\lambda_t^2 = \frac{1-\bar{\alpha}^{\prime}_{t-1}}{1-\bar{\alpha}_t^{\prime}}(1-\alpha_t^{\prime}),\quad \Gamma_n=\frac{1}{M} \sum_{m=1}^M \frac{\left\|\boldsymbol{s}_t\left(\boldsymbol{x}_{t, m}\right)\right\|^2}{d}, \boldsymbol{x}_{t, m} \stackrel{i.i.d}{\sim} q_t\left(\boldsymbol{x}_t\right)    
\end{equation}

The corresponding iteration rule is:
\begin{equation}
    \E(\vx_{t-1})=\frac{1}{\sqrt{\alpha^{\prime}_t}}\E(\vx_t)-\frac{1-\alpha^{\prime}_t}{\sqrt{\alpha^{\prime}_t(1-\bar{\alpha}^{\prime}_t)}}\E(\boldsymbol{\epsilon}_t)
\end{equation}
\begin{equation}
    \Var(\mathbf{x}_{t-1})=\frac{1}{\alpha^{\prime}_t}\Var(\mathbf{x}_t)-2\frac{1-\alpha^{\prime}_t}{\alpha^{\prime}_t\sqrt{1-\bar{\alpha}^{\prime}_t}}\Cov(\vx_t,\boldsymbol{\epsilon}_t)+\frac{(1-\alpha^{\prime}_t)^2}{\alpha^{\prime}_t(1-\bar{\alpha}^{\prime}_t)}\Var(\boldsymbol{\epsilon}_t)+\sigma_t^2
\end{equation}

For DDIM \citep{song2020denoising}, the sampling method is 
\begin{equation}
\vx_{t-1}=\alpha_{t-1}(\frac{\vx_t-\sigma_t\boldsymbol{\epsilon}_t}{\alpha_t})+\sigma_{t-1}\boldsymbol{\epsilon}_t  
\end{equation} 
The corresponding iteration rule is:
\begin{equation}
    \E(\vx_{t-1})=\frac{\alpha_{t-1}}{\alpha_t}\E(\vx_t)+(\sigma_{t-1}-\frac{\alpha_{t-1}}{\alpha_t}\sigma_{t})\E(\boldsymbol{\epsilon}_t)
\end{equation}
\begin{equation}
\Var(\vx_{t-1})=\frac{\alpha_{t-1}^2}{\alpha_t^2}\Var(\vx_t)+2\frac{\alpha_{t-1}}{\alpha_t}(\sigma_{t-1}-\frac{\alpha_{t-1}}{\alpha_t}\sigma_{t})\Cov(\vx_t,\boldsymbol{\epsilon}_t)+(\sigma_{t-1}-\frac{\alpha_{t-1}}{\alpha_t}\sigma_{t})^2\Var(\boldsymbol{\epsilon}_t)
\end{equation}

For 2-order DPM-Solver~\citep{lu2022dpm}, applying our method is non-trivial because it involves an extra hidden state between $\vx_t$ and $\vx_{t-1}$:
\begin{equation} 
\label{eq:dpm-solver}
\vx_{s_t}=\frac{\alpha_{s_t}}{\alpha_t} \vx_t-\sigma_{s_t}(e^{\frac{h_t}{2}}-1)\boldsymbol{\epsilon}_t
\end{equation}
\begin{equation}
\label{eq:dpm2-2}
\vx_{t-1}=\frac{\alpha_{t-1}}{\alpha_t} \vx_t-\sigma_{t-1}(e^{h_t}-1) \boldsymbol{\epsilon}_{s_t},
\end{equation}
where $\lambda_t=\log \frac{\alpha_t}{\sigma_t} $ is the half-log-SNR~\citep{lu2022dpm}. 
$ h_t=\lambda_{t-1}-\lambda_t.$
$s_t$ denotes the timestep corresponding to the half-log-SNR of $\frac{\lambda_{t-1}+\lambda_t}{2}$. 

Estimating expectation and variance for both sides of \cref{eq:dpm2-2} yields:
\begin{equation}
\E(\vx_{t-1})=\frac{\alpha_{t-1}}{\alpha_t} \E(\vx_t)-\sigma_{t-1}(e^{h_t}-1) \E(\boldsymbol{\epsilon}_{s_t})
\end{equation}
\begin{equation}
\small
\Var(\vx_{t-1})=\frac{\alpha_{t-1}^2}{\alpha_t^2} \Var(\vx_t)-2\frac{\alpha_{s_t}}{\alpha_t}\sigma_{t-1}(e^{h_t}-1)\Cov(\vx_t, \boldsymbol{\epsilon}_{s_t})+\sigma_{t-1}^2(e^{h_t}-1)^2\Var(\boldsymbol{\epsilon}_{s_t}).
\end{equation}
Unlike first-order sampling methods, we cannot approximate $\Cov(\vx_t, \boldsymbol{\epsilon}_{s_t})$ with rarely MC samples $\vx_{t,i}$. 
Nonetheless, we observe that
injecting the uncertainty on $\boldsymbol{\epsilon}_t$ to 
\cref{eq:dpm-solver} yields
\begin{equation}
\label{eq:u|x}
\vx_{s_t} | \vx_t \sim \mathcal{N}(\frac{\alpha_{s_t}}{\alpha_t} \vx_t-\sigma_{s_t}(e^{\frac{h_t}{2}}-1){\epsilon}_{\theta}(\vx_t, t), \mathrm{diag}(\sigma_{s_t}^2(e^{\frac{h_t}{2}}-1)^2{\gamma}^2_\theta(\vx_t, t))).
\end{equation}
Consequently, we can first sample $\vx_{t,i} \sim \mathcal{N}(\E(\vx_{t}), \Var(\vx_{t}))$, based on which $\vx_{s_t, i}$ are sampled. 
Then, we can approximate $\Cov(\vx_t, \boldsymbol{\epsilon}_{s_t})$ with MC estimation similar to \cref{eq:mc}.

\subsection{Uncertainty Quantification on Continuous-time Reverse Process}
\label{append:continuous}
Firstly, we integrate \cref{eq:sampling} with time $t$ from $0$ to $T$ to obtain the continuous-time solution $\vx_0$.
\begin{equation}
    x_0=x_T-\int_0^T [f(t)\vx_t + \frac{g(t)^2 }{\sigma_t}\boldsymbol{\epsilon}_t] dt + \int_0^T g(t)d \bar{\boldsymbol{\omega}}_t
\end{equation}
Due to the independence between the reverse-time Wiener process and $x_t$, we have
\begin{equation}
\begin{aligned}
    \Var(\vx_0) &= \Var(\vx_T)+\Var(\int_0^T [f(t)\vx_t + \frac{g(t)^2 }{\sigma_t}\boldsymbol{\epsilon}_t] dt)+\Var(\int_0^T g(t)d \bar{\boldsymbol{\omega}}_t) \\
    &= \mathbf{1}+\Var(\int_0^T [f(t)\vx_t + \frac{g(t)^2 }{\sigma_t}\boldsymbol{\epsilon}_t] dt) + \int_0^T g(t)^2 dt
\end{aligned}   
\end{equation}

The second equality is derived using the Ito isometry property \cref{eq:ito} of Ito calculus:
\begin{equation}
\label{eq:ito}
    \E(\int_0^T g_1(t)d \bar{\boldsymbol{\omega}}_t \int_0^T g_2(t)d \bar{\boldsymbol{\omega}}_t ) = \E(\int_0^T g_1(t)g_2(t) dt )
\end{equation}
that is, 
\begin{equation}
\begin{aligned}
  \Var(\int_0^T g(t)d \bar{\boldsymbol{\omega}}_t) = &\E((\int_0^T g(t)d \bar{\boldsymbol{\omega}}_t)^2)-\E(\int_0^T g(t)d \bar{\boldsymbol{\omega}}_t)^2 \\ = &\E((\int_0^T g(t)d \bar{\boldsymbol{\omega}}_t)^2)- 0 = \int_0^T g(t)^2 dt
\end{aligned}  
\end{equation}

Assuming that the reward process ${\vx_t, t \in [0,T]}$ is a stochastic process with second order moments and is mean square integrable, we have
\begin{equation}
\begin{aligned}
    & \Var(\int_0^T [f(t)\vx_t + \frac{g(t)^2 }{\sigma_t}\boldsymbol{\epsilon}_t] dt) \\
    =& \int_0^T\int_0^T f(s)f(t)\Cov (\vx_s,\vx_t) \\
    -& f(s)\frac{g(t)^2 }{\sigma_t}\Cov (\vx_s, \boldsymbol{\epsilon}_t) - f(t)\frac{g(s)^2 }{\sigma_s}\Cov (\vx_t, \boldsymbol{\epsilon}_s) + \frac{g(s)^2 }{\sigma_s}\frac{g(t)^2 }{\sigma_t}\Cov (\boldsymbol{\epsilon}_s, \boldsymbol{\epsilon}_t) ds dt \\
    =& \int_0^T\int_0^T f(s)f(t)\Cov (\vx_s,\vx_t) - f(s)\frac{g(t)^2 }{\sigma_t}\Cov (\vx_s, \boldsymbol{\epsilon}_t) ds dt \\ 
    -& 2 \int_0^T f(s)\frac{g(s)^2 }{\sigma_s}\Cov (\vx_s, \boldsymbol{\epsilon}_s) ds + \int_0^T \frac{g(s)^4 }{\sigma_s^2}\Cov (\boldsymbol{\epsilon}_s, \boldsymbol{\epsilon}_s) ds
\end{aligned}   
\end{equation}

The second equvalence holds because $\vx_s$ and $\boldsymbol{\epsilon}_t$,  $\boldsymbol{\epsilon}_s$ and $\boldsymbol{\epsilon}_t$ are both independent when $s \neq t$, i.e. $\Cov (\vx_s, \boldsymbol{\epsilon}_t)=0, \Cov (\boldsymbol{\epsilon}_s, \boldsymbol{\epsilon}_t)=0, \forall s \neq t$.
\begin{equation}
\begin{aligned}
    \int_0^T\int_0^T f(s)f(t)\Cov (\vx_s,\vx_t) ds dt =&
    2 \int_0^T f(t)(\int_0^t f(s)\Cov(\vx_s, \vx_t)ds) dt \\
    =& 2\int_0^T f(t)(\int_0^t f(s)\Cov(\vx_s, \vx_s + \int_s^t (f(u)\vx_u-\frac{g(u)^2 }{\sigma_u}\boldsymbol{\epsilon}_u) du \\
    +&\int_s^t g(u) d \bar{\boldsymbol{\omega}}_u ds) dt \\
    =& 2 \int_0^T f(t)(\int_0^t f(s)\Var(\vx_s) ds) dt \\+&
    2\int_0^T f(t)(\int_0^t f(s)\Cov(\vx_s, \int_s^t (f(u)\vx_u-\frac{g(u)^2}{\sigma_u}\boldsymbol{\epsilon}_u) du) ds) dt 
\end{aligned} 
\end{equation}

Then the question boils down to approximate $\Cov(\vx_s, \int_s^t (f(u)\vx_u-\frac{g(u)^2}{\sigma_u}\boldsymbol{\epsilon}_u) du)$. 
According to the numerical integration method and $\Cov (\vx_s, \boldsymbol{\epsilon}_t)=0, \forall s \neq t$,
\begin{equation}
\begin{aligned}
    \Cov(\vx_s, \int_s^t (f(u)\vx_u-\frac{g(u)^2}{\sigma_u}\boldsymbol{\epsilon}_u) du) &= \Cov(\vx_s, \int_s^t f(u)\vx_u du) +\Cov(\vx_s, \int_s^t-\frac{g(u)^2}{\sigma_u}\boldsymbol{\epsilon}_u) du ) \\
    &=\Cov(\vx_s, \int_s^t f(u)\vx_u du)-
\frac{g(u)^2}{\sigma_u}\Cov(\vx_s,\boldsymbol{\epsilon}_u) \Delta t
\end{aligned}
\end{equation}

 So when $\Delta  t \rightarrow 0$, 
 \begin{equation}
\Cov(\vx_s, \int_s^t (f(u)\vx_u-\frac{g(u)^2}{\sigma_u}\boldsymbol{\epsilon}_u) du) = \Cov(\vx_s, \int_s^t f(u)\vx_u du)
 \end{equation}

According to sampling method $\vx_{t-1}=\vx_{t}-(f(t)\vx_t+\frac{g(t)^2}{\sigma_t}\boldsymbol{\epsilon}_t))\Delta t+g(t)(\vw_{t-1}-\vw_{t})$, 
\begin{equation}
\begin{aligned}
    &x_{s+\Delta t}=x_s+(f(s)\vx_s+\frac{g(s)^2}{\sigma_s}\boldsymbol{\epsilon}_s)\Delta t+h(w)\\
    &x_{s+2\Delta t}=x_{s+\Delta t}+(f(s+\Delta t)\vx_{s+\Delta t}+\frac{g(s+\Delta t)^2}{\sigma_{t+\Delta t}}\boldsymbol{\epsilon}_{s+\Delta t})\Delta t+h(w) \\
    &\qquad =x_s+(f(s)\vx_s+\frac{g(s)^2}{\sigma_s}\boldsymbol{\epsilon}_s)\Delta t+ 
    f(s+\Delta t)\vx_s\Delta t+\frac{g(s+\Delta t)^2}{\sigma_{t+\Delta t}}\boldsymbol{\epsilon}_{s+\Delta t}\Delta t+ \mathcal{O}(\Delta t^2)+h(w)
\end{aligned}     
\end{equation}

Then using naiive numerical integration method, we have
\begin{equation}
\begin{aligned}
    \int_s^t f(u)\vx_u du 
    &=f(s)\vx_s \Delta t+f(s+\Delta t)\vx_{s+\Delta t} \Delta t+f(s+2\Delta t)\vx_{s+2\Delta t} \Delta t+\dots++f(t-\Delta t)\vx_{t-\Delta t} \Delta t\\
    &= f(s)\vx_s \Delta t+f(s+\Delta t)\vx_s \Delta t+f(s+2\Delta t)\vx_s \Delta t+ \dots +f(t-\Delta t)\vx_s \Delta t +\sum \mathcal{O}(\Delta t^2)+H(w)\\
    &= \int_s^t f(u)\vx_s du+\sum \mathcal{O}(\Delta t^2)+H(w)
\end{aligned}   
\end{equation}
We neglect second-order terms and get the approximation of $\Cov(\vx_s, \int_s^t f(u)\vx_u du)$:
\begin{equation}
    \Cov(\vx_s, \int_s^t f(u)\vx_u du)\approx \Cov(\vx_s, \int_s^t f(u)\vx_s du)= \Var(\vx_s)\int_s^t f(u)du
\end{equation}

In conclusion, we derive an approximate illustrating the pattern of uncertainty dynamics from $\vx_T$ to $\vx_0$, 
\begin{equation}
\begin{aligned}
    \Var(\vx_0) \approx \mathbf{1} &+
    2 \int_0^T f(t)(\int_0^t f(s)\Var(\vx_s) ds) dt +2 \int_0^T f(t)(\int_0^t f(s) \Var(\vx_s)(\int_s^t f(u)du) ds) dt\\
    & - 2 \int_0^T f(s)\frac{g(s)^2 }{\sigma_s}\Cov (\vx_s, \boldsymbol{\epsilon}_s) ds + \int_0^T \frac{g(s)^4 }{\sigma_s^2}\Var (\boldsymbol{\epsilon}_s) ds + \int_0^T g(t)^2 dt
\end{aligned}   
\end{equation}
Moreover, we can generalize it to arbitrary reverse time interval $[i, j] \in [0, T]$. 

Specifically, $\forall  0\leq i \leq j \leq T $, 
\begin{equation}
\begin{aligned}
    \Var(\vx_i) \approx \Var (\vx_{j}) &+
    2 \int_i^{j} f(t)(\int_i^t f(s)\Var(\vx_s) ds) dt+2 \int_i^{j} f(t)(\int_i^t f(s) \Var(\vx_s)(\int_s^t f(u)du) ds) dt \\
    & - 2 \int_i^{j} f(s)\frac{g(s)^2 }{\sigma_s}\Cov (\vx_s, \boldsymbol{\epsilon}_s) ds + \int_i^{j} \frac{g(s)^4 }{\sigma_s^2}\Var (\boldsymbol{\epsilon}_s) ds + \int_i^{j} g(t)^2 dt
\end{aligned}  
\end{equation}

\subsection{BayesDiff-Skip Alogorithm}
In this section, we present our BayesDiff-Skip algorithm. To be specific, if sampling with uncertainty is used from $\vx_{t}$ to $\vx_{t-1}$, then $\boldsymbol{\epsilon}_{t}$ is considered as a random variable sampled from the normal posterior predictive distribution, where $\Cov(\vx_t, \boldsymbol{\epsilon}_{t})$ and $\Var(\boldsymbol{\epsilon}_{t})$ are non-zero. Conversely, if original deterministic sampling is used from $\vx_{t}$ to $\vx_{t-1}$, $\boldsymbol{\epsilon}_{t}$ is treated as a constant and $\Cov(\vx_t, \boldsymbol{\epsilon}_{t})$ and $\Var(\boldsymbol{\epsilon}_{t})$ are zero. We conclude it to this algorithm in \cref{algo:skipuq}. 
\begin{algorithm}[t]
    \caption{A faster variant of BayesDiff. (BayesDiff-Skip)}
    \label{algo:skipuq}
    \begin{algorithmic}[1]  
        \Require Starting point $\vx_T$, Monte Carlo sample size $S$, Pre-trained noise prediction model $\epsilon_\theta$.
        \Ensure  Image generation $\vx_0$ and pixel-wise uncertainty $\Var(\vx_0)$.
        \State Construct the pixel-wise variance prediction function ${\gamma}^2_\theta$ via LLLA;
        \State $\E(\vx_T)\gets\vx_T, \Var(\vx_T)\gets\mathbf{0}$, $\Cov(\vx_T, \boldsymbol{\epsilon}_T)\gets\mathbf{0}$;
        \For {$t=T \to 1$}
            \If{$t \in \tilde{\vt}$}
                \State Sample $\boldsymbol{\epsilon}_{t} \sim \mathcal{N}({\epsilon}_{\theta}(\vx_t, t), \mathrm{diag}({\gamma}^2_\theta(\vx_t, t)))$;
            \Else
                \State $\boldsymbol{\epsilon}_{t} \gets {\epsilon}_{\theta}(\vx_t, t)$, $\Cov(\vx_t,\boldsymbol{\epsilon}_t) \gets \mathbf{0}$, $\Var(\boldsymbol{\epsilon}_{t}) \gets \mathbf{0}$;
            \EndIf
            \State Obtain $\vx_{t-1} $ via \cref{eq:disc};
            \State Estimate $\E(\vx_{t-1})$ and $\Var(\vx_{t-1})$ via \cref{eq:exp-sde} and \cref{eq:var-ite};
            \If{$t-1 \in \tilde{\vt}$}
                \State Sample $\vx_{t-1,i} \sim \mathcal{N}(\E(\vx_{t-1}), \Var(\vx_{t-1})), i=1,\dots, S$;
                \State Estimate $\Cov(\vx_{t-1},\boldsymbol{\epsilon}_{t-1}) $ via \cref{eq:mc};
            \EndIf
        \EndFor
    \end{algorithmic}
\end{algorithm}
\label{append:bayesdiff}

\subsection{Implementation Details of Last Layer Laplace Approximation}
\label{llla_implement_details}
We adopt the most lightweight diagonal factorization and ignore off-diagonal elements for Hessian approximation in LLLA~\citep{daxberger2021laplace}. To avoid storing large Jacobian matrix, we adopt the Monte Carlo method to approximate the accurate variance of outputs $diag(\gamma_\theta^2(x, t))$ directly by the variance of samples { $ f_{\theta_i}(x,t) $ }, $\theta_i  \sim p(\theta|\mathcal{D}) =\mathcal{N}(\theta ; \theta_{\text{MAP}}$, $\boldsymbol{\Sigma})$. This results in faster computation speed, while still maintaining a reasonable level of accuracy. The number of samples is chosen as 100 in practice. 
\subsection{Additional Examples of Resampling Method with BayesDiff}
\label{append:more_examples}
\begin{figure}[t]
\begin{center}
\includegraphics[width=0.9\textwidth]{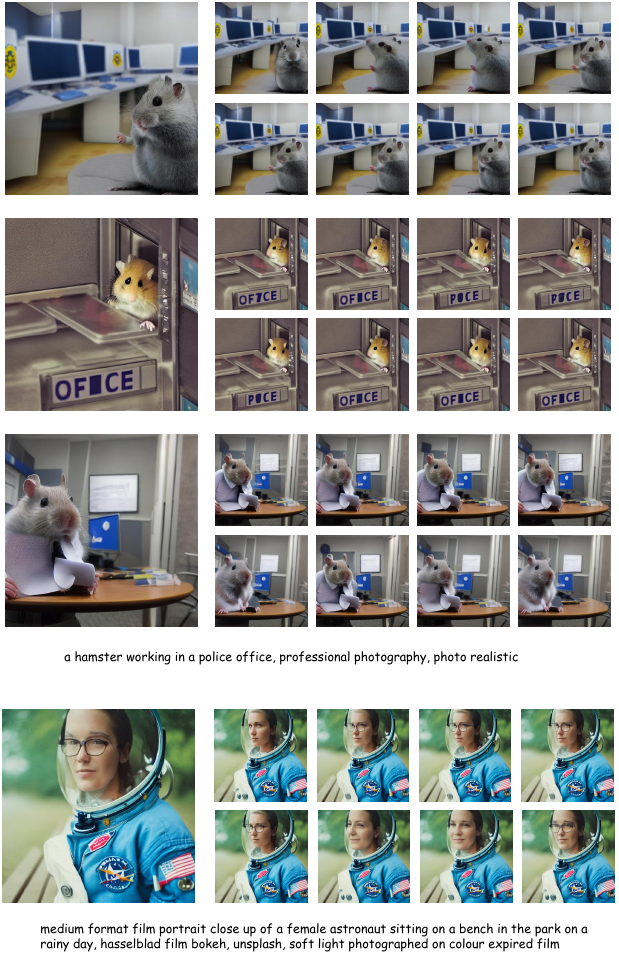}
\end{center}
\caption{8 potential rectified samples resampled by BayesDiff (right) of flawed images (left) annotated by humans. }
\label{fig:rec_1}
\end{figure}

\begin{figure}[t]
\begin{center}
\includegraphics[width=0.9\textwidth]{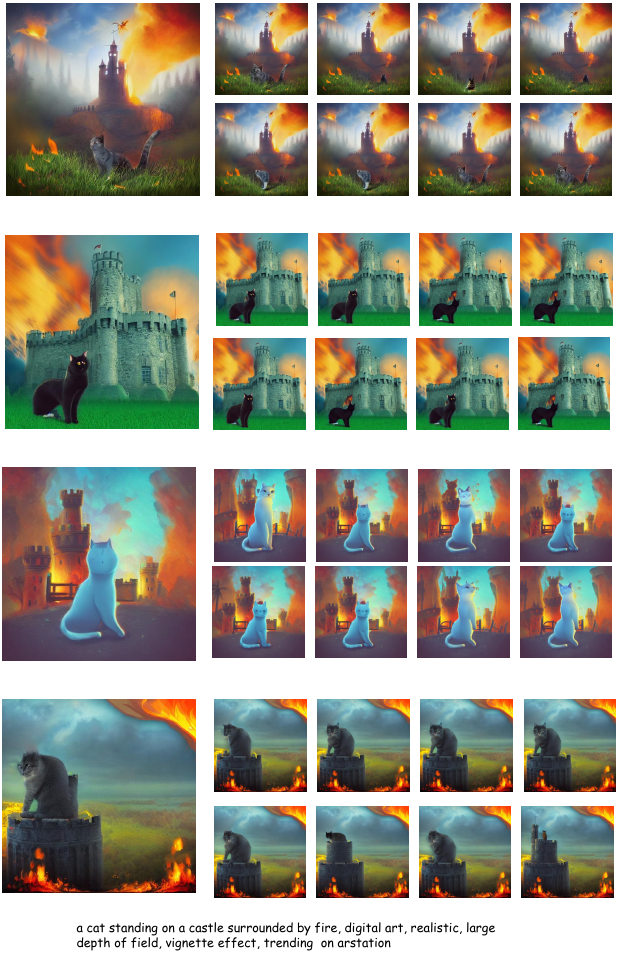}
\end{center}
\caption{8 potential rectified samples resampled by BayesDiff (right) of flawed images (left) annotated by humans.}
\label{fig:rec_2}
\end{figure}

To provide a more intuitive demonstration of the significance of our resampling method, we include additional prompts and corresponding images containing artifacts annotated by humans, presenting 8 resampled images using BayesDiff in \cref{fig:rec_1} and \cref{fig:rec_2}. We find that the success rate of resampling flawed samples into desirable samples is approximately at least $60\%$.
\end{document}